\documentclass{article}

\usepackage[preprint]{neurips_2025}

\usepackage{threeparttable}

\usepackage[utf8]{inputenc} 
\usepackage[T1]{fontenc}    
\usepackage{hyperref}       
\usepackage{url}            
\usepackage{booktabs}       
\usepackage{amsfonts}       
\usepackage{nicefrac}       
\usepackage{microtype}      
\usepackage{xcolor}         

\usepackage{microtype}
\usepackage{graphicx}
\usepackage{subfigure}
\usepackage{booktabs} 
\usepackage{caption}

\usepackage{hyperref}

\usepackage{amsmath}
\usepackage{amssymb}
\usepackage{mathtools}
\usepackage{amsthm}
\usepackage[misc]{ifsym}
\usepackage[capitalize,noabbrev]{cleveref}

\theoremstyle{plain}

\theoremstyle{definition}

\theoremstyle{remark}

\usepackage[textsize=tiny]{todonotes}
\usepackage{newfloat}
\usepackage{enumitem}
\usepackage{listings}
\usepackage{tcolorbox}
\usepackage{lipsum}
\usepackage{listings}
\usepackage{courier}

\usepackage{amsmath,amsfonts,bm}

\def\eqref#1{equation~\ref{#1}}

\def\1{\bm{1}}

\DeclareMathAlphabet{\mathsfit}{\encodingdefault}{\sfdefault}{m}{sl}
\SetMathAlphabet{\mathsfit}{bold}{\encodingdefault}{\sfdefault}{bx}{n}

\def\gD{{\mathcal{D}}}

\def\gG{{\mathcal{G}}}

\def\gI{{\mathcal{I}}}

\def\gS{{\mathcal{S}}}

\usepackage{multirow}
\usepackage{makecell}
\usepackage{amsmath}
\tcbset{
  colback=gray!10, 
  colframe=black, 
  fonttitle=\bfseries, 
  sharp corners, 
  boxrule=0.4mm, 
  top=2mm, bottom=2mm, left=6mm, right=0mm, 
  boxsep=0mm, 
  width=\textwidth, 
}

\title{Hierarchical Instruction-aware \\Embodied Visual Tracking}

\author{
  \textbf{Kui Wu$^1$}, \textbf{Hao Chen$^2$}, \textbf{Churan Wang$^3$}, \textbf{ Fakhri Karray$^4$},\\
  \textbf{Zhoujun Li$^1$}, \textbf{Yizhou Wang$^{3}$}, \textbf{Fangwei Zhong$^{5,\textrm{\Letter}}$}\\
  $^1$Beihang University,
  $^2$City University of Macau,
  $^3$Peking University,\\
  $^4$Mohamed bin Zayed University of Artificial Intelligence,
  $^5$Beijing Normal University\\
  \texttt{wukui0099@gmail.com}, \texttt{fangweizhong@bnu.edu.cn}\\
}

\begin{document}
\maketitle
\renewcommand{\thefootnote}{} 
\footnotetext{$\textrm{\Letter}$ Corresponding authors.}

\begin{figure*}[h]
    \centering
    \vspace{-0.6cm}
    \includegraphics[width=0.99\textwidth]{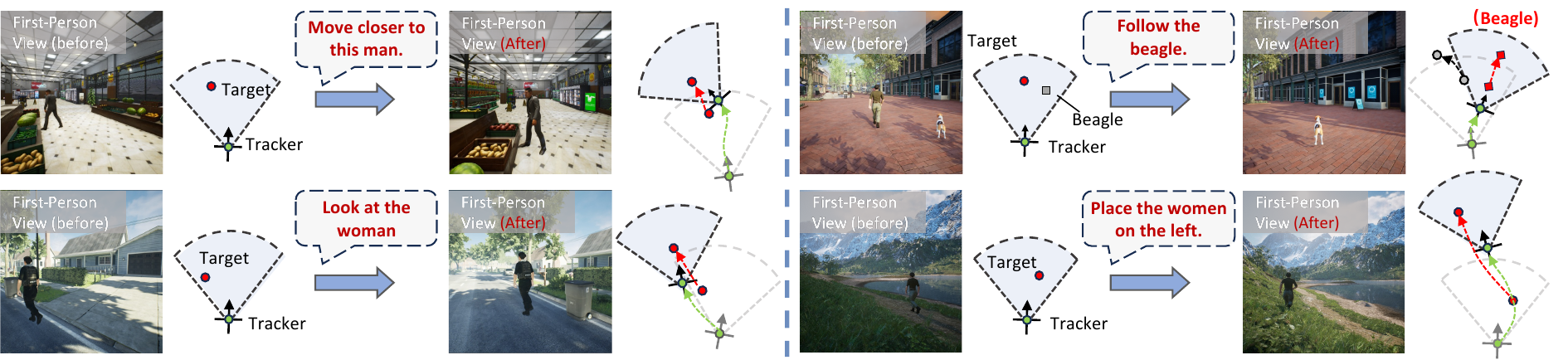}
     \caption{Examples of User-Centric embodied visual tracking with diverse instructions.}
    \label{fig:instruct_evt}
\end{figure*}
\begin{abstract}

User-Centric Embodied Visual Tracking (UC-EVT) presents a novel challenge for reinforcement learning-based models due to the substantial gap between high-level user instructions and low-level agent actions. While recent advancements in language models (e.g., LLMs, VLMs, VLAs) have improved instruction comprehension, these models face critical limitations in either inference speed (LLMs, VLMs) or generalizability (VLAs) for UC-EVT tasks. To address these challenges, we propose \textbf{Hierarchical Instruction-aware Embodied Visual Tracking (HIEVT)} agent, which bridges instruction comprehension and action generation using \textit{spatial goals} as intermediaries. HIEVT first introduces \textit{LLM-based Semantic-Spatial Goal Aligner} to translate diverse human instructions into spatial goals that directly annotate the desired spatial position. Then the \textit{RL-based Adaptive Goal-Aligned Policy}, a general offline policy, enables the tracker to position the target as specified by the spatial goal. To benchmark UC-EVT tasks, we collect over ten million trajectories for training and evaluate across one seen environment and nine unseen challenging environments. Extensive experiments and real-world deployments demonstrate the robustness and generalizability of HIEVT across diverse environments, varying target dynamics, and complex instruction combinations. The complete project is available at \url{https://sites.google.com/view/hievt}.
\end{abstract}

\section{Introduction}

Providing \textbf{User-Centric Embodied Visual Tracking}~(UC-EVT) has become increasingly appealing for modern intelligent agents, as users demand more dynamic and interactive systems beyond fixed targets and distance~\cite{van2023human,zhang2023animaltrack}. In particular, users expect systems that can quickly comprehend instructions and respond effectively to dynamic and complex tracking scenarios~\cite{li2023exploring,zhou2024hazard,ma2023chimpact}. Motivated by the above needs, this paper proposes three core requirements for UC-EVT:
\textbf{1) User Instructions Understanding.} The UC-EVT agent must be capable of interpreting user instructions, such as natural language commands, to dynamically adjust the tracking agent, including targets, angles, and distances.
\textbf{2) Real-Time Responsiveness.} The agent must operate in real time, maintaining robust performance by efficiently tracking fast-moving or unseen targets in dynamic environments.
\textbf{3) Flexibility and Generalizability.} The agent should flexibly adapt to various targets, changing environments, diverse instructions, and evolving spatial relationships, all without requiring retraining.

Existing methods for embodied visual tracking (EVT)~\cite{luo2018journal, zhong2019ad, zhong2021distractor, zhong2023rspt} primarily focus on static or predefined goals, such as maintaining fixed distances or angles relative to a target. Reinforcement learning (RL)-based models, for example, utilize the distance from the target to the expected location~(usually at the center of the view) as the reward to train tracking policy in an end-to-end manner~\cite{luo2018journal, zhong2019ad, zhong2021distractor} or based on vision foundation models~\cite{zhong2023rspt, zhong2024empowering}. 
Additionally, systems like Vision-Language-Action Models (VLA) and general-purpose large models~(e.g., LLMs and VLMs) have demonstrated improved instruction comprehension capabilities due to their large-scale parameterization.

Despite these advancements, existing EVT models face significant limitations when applied to UC-EVT tasks. Specifically:
\underline{1) Limited Comprehension and Flexibility in RL-based EVT:} RL-based EVT models struggle to handle complex user instructions and exhibit poor transferability across different environments and embodiments. Their rigid tracking strategies (e.g., fixed distances or angles) hinder adaptability to user-centric instructions.
\underline{2) Limited Generalization of VLA Models:} VLA models rely heavily on large-scale annotated data, making them ill-suited for unseen conditions such as novel environments, instructions, targets, or agents.
\underline{3) Inference Latency in Large Models:} While large models exhibit strong capabilities, their inference speeds (typically 0.5--3 FPS) are insufficient for real-time tracking, resulting in frequent target loss during fast movements.

To address these limitations, we propose a novel \textbf{Hierarchical Instruction-aware Embodied Visual Tracking (HIEVT)} agent that integrates instruction comprehension models with adaptive tracking policies and introduces intermediate spatial goals to bridge human instructions and agent behavior. Specifically, HIEVT introduces an \textit{LLM-based Semantic-Spatial Goal Aligner}, which translates diverse human instructions into spatial goals. This is followed by the \textit{RL-Based Adaptive Goal-Aligned Policy}, a general offline policy enabling the agent to efficiently approximate the spatial goal and achieve precise tracking.

Our contributions are summarized as follows:
1) We introduce the User-Centric Embodied Visual Tracking (UC-EVT) task, which lays the foundation for user-centric human-robot interactions, enabling robots to follow users and provide personalized services. 
2) We propose a novel Hierarchical Instruction-aware EVT (HIEVT) model that effectively addresses the limitations of state-of-the-art (SOTA) models while preserving their advantages.
3) We benchmark the UC-EVT task by preparing a large-scale dataset of 10 million annotated trajectories and conducting extensive evaluations across 10 diverse virtual environments. Additionally, we implement RL-based, VLA, and VLM baselines for the UC-EVT task, with all codes and datasets made publicly available.
4)  Our extensive experiments across 10 virtual environments and different moving speeds demonstrate that HIEVT outperforms existing baselines significantly. Furthermore, real-world deployments in three different environments validate the robustness and effectiveness of our approach.

\vspace{-0.2cm}
\section{Preliminaries}
\vspace{-0.1cm}

\vspace{-0.2cm}
\paragraph{Problem Definition.} User-Centric Embodied Visual Tracking (UC-EVT)  starts with an initial user instruction and a target. The tracker then attempts to track the target immediately by following the user's instructions. The main components of this problem are outlined as follows:

\begin{itemize}[leftmargin=*]
    \item \textbf{User Instruction.} The user initially provides an instruction $\mathcal{I}_0$, which serves as the starting command for the tracking agent. At any subsequent time step $t$, the user can issue a new instruction $\mathcal{I}_t$ to update the agent's behavior. These instructions can be provided in natural language (e.g., ``Move closer'') or by directly specifying a location, such as drawing a bounding box.
    \item \textbf{Tracking State.} The tracking agent's state at time $t$, denoted as $s_t$, is represented by the relative distance $\rho_t$ and the relative angle $\theta_t$ with respect to the target. Specifically, the tracking state is given as:
    $\mathcal{S}_t = (\rho_t, \theta_t)$.
    \item \textbf{Agent Action.} At each time step $t$, the tracking agent executes an action $a_t$ to adjust its position and orientation in order to minimize the discrepancy between the current tracking state and the user's instruction.

    \item \textbf{Environment Transferring.} To ensure applicability to real-world scenarios, the tracker should be tested in unseen environments. This setup is designed to test the generalization and transferability of the tracking system.
\end{itemize}

\paragraph{Objective Function.} The goal of UC-EVT  is to minimize the discrepancy between the user instruction $\mathcal{I}_t$ and the tracking state $s_t$ throughout the entire tracking period $T$. This discrepancy is measured using a distance function $\mathcal{D}(\mathcal{I}_t, s_t)$. The objective function is defined as $\min \sum_{t=1}^{T} \mathcal{D}(\mathcal{I}_t, s_t)$,
where $\mathcal{D}(\mathcal{I}_t, s_t)$ quantifies the difference between the user instruction $\mathcal{I}_t$ and the tracking state $s_t$ at each time step $t$. The objective aims to ensure that the tracking agent follows the user's instructions as closely as possible over the entire tracking period (episode).
\section{Hierarchical Instruction-aware Embodied Visual Tracking}
\label{method}
\begin{figure}
    \centering
    \includegraphics[width=0.9\linewidth]{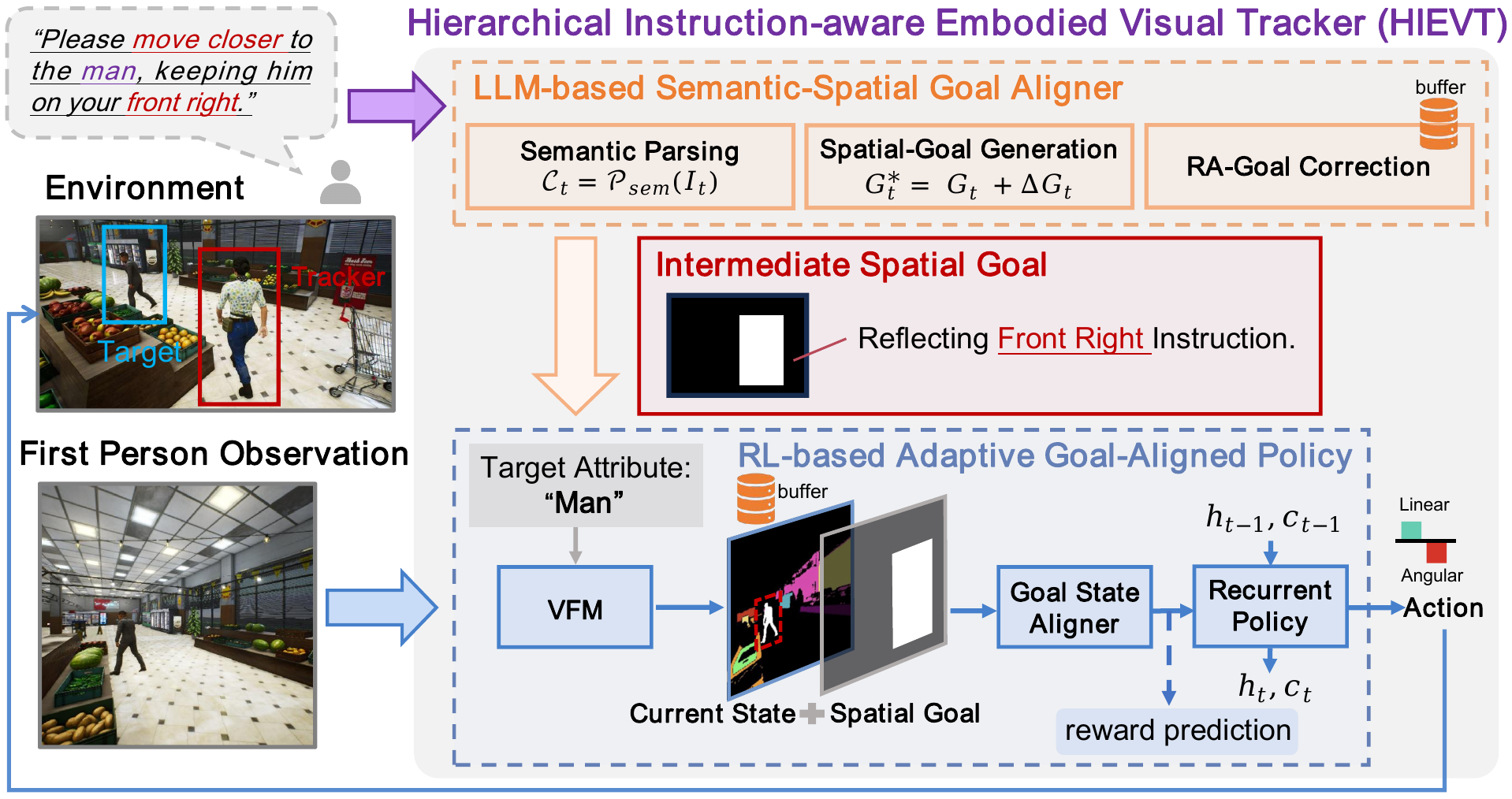}
    \caption{Overview of the Hierarchical Instruction-aware Embodied Visual Tracker (HIEVT). Given a natural language instruction and environmental observation, our system first processes the instruction through the LLM-based Semantic-Spatial Goal Aligner including Semantic Parsing, Spatial-Goal Generation, and Retrieval-Augmented Goal Correction. This produces a target attribute and a bounding box format spatial goal. The RL-based Adaptive Goal-Aligned Policy then combines this goal with the Visual Foundation Model (VFM) processed observation, feeds them into the following policy network. The Goal State Aligner and Recurrent Policy then generate appropriate action signals to maintain the desired spatial relationship with the target.}
    \label{fig:policy_input}
    \vspace{-0.5cm}
\end{figure}

In this section, we present the design insight of our proposed model, \textbf{Hierarchical Instruction-aware Embodied Visual Tracking (HIEVT) }.
The core challenge of the User-Centric Embodied Visual Tracking (UC-EVT) is the giant gap from the user instruction $\gI_t$ and the actual state $\gS_t$ of the tracker. Moreover, the instruction $\gI_t$ is given by a user-friendly mode rather than an agent-friendly mode. To bridge the gap, it is necessary to import an intermediate goal $\gG_{inter}{(t)}$ to bridge the user's instruction and the agent's state. This decomposition allows for user instruction understanding as well as efficient agent decision-making, formulated as:
\begin{equation}
    \gD(\gI_t, \gS_t)\  \approx \ 
    \underbrace{\gD(\gI_t, \gG_{inter}(t))}_{
    \text{Semantic-Spatial Goal Aligner}
    }
    + \underbrace{\gD(\gG_{inter}(t),\gS_t)}_{
    \text{Adaptive Goal-Aligned Policy}
    },
    \label{eq:dual_goal_aligning}
\end{equation}
where $\gD(\gI_t, \gG_{inter}(t))$ represents the distance between the user instruction and the intermediate goal, and $\gD(\gG_{inter}(t),\gS_t)) $ represents the distance between the intermediate goal and the agent state. We then detail the design of the key components, LLM-based Semantic-Spatial Goal Aligner and RL-based Adaptive Goal-Aligned Policy, the entire model structure is illustrated in Figure ~\ref{fig:policy_input}.

\subsection{LLM-based Semantic-Spatial Goal Aligner}
The Semantic-Spatial Goal Aligner (SSGA) is the fundamental component responsible for translating the user instruction $\gI_t$ to a spatial goal $\gG_{inter}(t)$. Given the input user instruction $\gI_t$, the SSGA outputs an intermediate goal $\gG_{inter}(t) = (\mathcal{C}_t, G
_t)$, where $\mathcal{C}_t$ is the target category, and $G^*_t = [x_t, y_t, w_t, h_t]$ is a spatial goal, representing the expected target's spatial position in the bounding box format. The SSGA consists of three core components: semantic parsing, spatial-goal generation, and retrieval-augmented goal correction.

\paragraph{Semantic Parsing} 
The first step in SSGA is to interpret the user instruction $\gI_t$ and extract critical information for goal specification: the target category $\mathcal{C}_t$. This process is performed by the semantic parser  $\mathcal{C}_t = \mathcal{P}_{sem}(\gI_t)$,

where $\mathcal{C}_t \in \mathcal{T}$ indicates the target attributes. For instance, given the instruction ``Get closer to the blue car'' the semantic parser identifies $\mathcal{C}_t = \{\text{``blue"}, \text{``car"}\}$.

\paragraph{Spatial-Goal Generation} 
To resolve ambiguities in user instructions, our parser grounds its interpretation of input instruction $\gI_t$ in both linguistic context and visual scene features. This dual-grounding approach enables robust semantic alignment between natural language commands and environmental observations. Through this process, the parser generates a spatial goal representation $G^*_t$ that accurately corresponds to the intended spatial directives from instruction. This is achieved using a chain-of-thought (COT)-based reasoning mechanism ~\cite{wei2022chain}, which incrementally adjusts the current target's spatial representation $G_t$. The spatial goal generation process is modeled as:
\begin{equation}
G^*_t = G_t + \Delta G_t,  G_t=\mathcal{B}(\mathcal{VFM}(\mathcal{O}_t,\mathcal{C}_t))
\label{eq:bbox_generation}
\end{equation}

where $\mathcal{VFM}(\mathcal{O}_t,\mathcal{C}_t)$ uses Vision Foundation Models (VFMs) to extract text-conditioned segmentation mask with target-highlighted format given the target category name ~\cite{zhong2024empowering}, $\mathcal{B(\cdot)}$ detects target's bounding box from segmentation mask by filtering the target's corresponding mask color (white color), $\Delta G_t = [\Delta x_t, \Delta y_t, \Delta w_t, \Delta h_t]$ represents the adjustment to the current target's spatial representation. The adjustment is determined by
$\Delta G_{t} = \mathcal{F}_{COT}(\mathcal{\gI}_t, G_t)$,
where $\mathcal{F}_{COT}$ represents a chain-of-thought reasoning process implemented through a large language models. This reasoning process interprets how the spatial directive in instruction $\mathcal{\gI}_t$ should modify the current observed target's spatial representation $G_t$, producing adjustment parameters $\Delta G_t$ for both position and size. To guide this interpretation, we developed a system prompt that structures the model's reasoning, ensuring it correctly analyzes spatial relationships (provided in supplementary materials) and translates natural language instructions into geometric transformations. For example, if the Instruction is ``get closer'' the bounding box is scaled up by generating a positive change on width $\Delta w_t$, height $\Delta h_t$ and a negative vertical change $\Delta y_t$ due to the perspective effect. Similarly, if the directive is ``move to the left'', resulting in a negative value on the horizontal coordinate change $\Delta x_t$. The COT-based approach ensures that the bounding box adjustments are interpretable and aligned with the user’s intent, allowing for dynamic and context-aware reasoning about the spatial relationship between the tracker and the target.

\paragraph{Retrieval-Augmented Goal Correction}
After generating the spatial goal $G^*_t$, the SSGA applies a correction step using a retrieval-augmented generation (RAG) mechanism. This step ensures that the generated bounding box is consistent with trajectory priors stored in a memory bank $\mathcal{M}$. The RAG module retrieves similar instructions and their corresponding bounding boxes from $\mathcal{M}$, i.e.,
$G_{retr} = \mathcal{R}(\gI_t, \mathcal{M})$,
where $\mathcal{R}$ is the retrieval function (i.e. Cosine similarity). The retrieved spatial goal $G_{retr}$ is then used for determining if the generated goal $G^*_t$ satisfies the target's physical constraints (e.g., aspect ratio or perspective effect). If conflicts arise, specifically, the Intersection over Union (IoU) ~\cite{leal2015motchallenge,cordts2016cityscapes} value is lower than a threshold, the historical mask is used as the final spatial goal $G^*_t$:
\begin{equation}
G^*_t=\left\{\begin{array}{ll}
G^*_t, & \text {IoU}\left(G^*_t, G_{retr}\right)>0.5 \\
G_{ {retr }}, & \text { IoU }\left(G^*_t, G_{ {retr }}\right) \leqslant 0.5
\end{array}\right.
\label{eq:fusion}
\end{equation}
This correction mechanism combines the adaptability of COT-based reasoning with the consistency of target physical constraints, ensuring robust and accurate bounding box predictions.

\subsection{RL-based Adaptive Goal-Aligned Policy}
\label{sec:goal_oriented_evt}

The proposed Adaptive Goal-Aligned Policy (AGAP) module bridges the intermediate goal $\mathcal{G}_{inter}(t)$ and the user implied state $\mathcal{S}_t$, ensuring precise alignment of the agent's actions with the desired user instructions. This alignment is achieved by an adaptive motion policy which dynamically adjust the agent's movement based on the observation $\mathcal{O}_t$ towards the spatial position indicated by $\mathcal{G}_{inter}(t)$. The adaptive policy is optimized using offline reinforcement learning with an auxiliary reward regression task to facilitate the alignment, formulated as: $\pi(a_t \mid G^*, \mathcal{O}_t)$. Below, we elaborate on the key components of this module.

\paragraph{Architechture}
The policy consists of two main components: 1) \textbf{Goal-State Aligner} contains multiple convolutional neural network (CNN) layers, serving as the feature extraction and alignment module, encoding the VFM pre-processed observations and the spatial goal representation at image-level space. This aligner outputs a latent aligned representation. Then, a reward prediction layer follows the aligner, serving as the auxiliary task to improve the aligned ability in a self-supervised manner. 2) \textbf{Recurrent Policy Network} consists of a long short-term memory (LSTM) network that models the temporal dynamics of the tracking process to enhance the spatial-temporal consistency of the representation and an actor network $V_{\phi}$. The latent representation from the Goal-state aligner was fed to the LSTM network, followed by the Actor Network to generate motion control action $a_t$. More details about the networks are in supplementary materials.  

At inference time, our system maintains \textbf{asynchronous processing} between goal generation and policy execution. Due to variable latency in LLM inference or network delays, spatial goals are not generated at fixed intervals. Instead, the goal-state aligner always uses the most recently available goal representation $G^*$, ensuring the tracking policy continues functioning smoothly without waiting for new goal updates. 
The inputs to the policy at any time step $t$ consist of: 1) The latest available spatial representation $G^*$, representing the expected target's spatial position at the image level, and 2) The VFM-processed tracker's first-person view observations $\mathcal{VFM}(\mathcal{O}_t,\mathcal{C}_t)$, as shown in Figure~\ref{fig:policy_input}. The low-level policy outputs an action $a_t$ that adjusts the tracker's movement, minimizing the discrepancy between the target position in observation $\mathcal{O}_{t}$ and the inferred spatial goal $G^*_t$.

\paragraph{Goal-conditioned Offline Policy Optimization}
For Adaptive Goal-Aligned Policy (AGAP), our aim is to train a policy that can adapt to diverse spatial goals and achieve precise and fast alignment. Traditional Online RL methods are typically designed for a single-goal objective, and require a huge amount of trial-and-error to converge. Therefore, we use the offline reinforcement learning (Offline-RL) paradigm to train the goal-conditioned policy, considering the training efficiency and dataset diversity's contribution to the overall performance. Meanwhile, we introduce an auxiliary regression task to improve the alignment ability. Below, we detail the reward design, offline-RL training objectives and the auxiliary regression task.

\begin{itemize}[leftmargin=*]{
\item \underline{Training Data Preparation.}
We extend the data collection procedure in ~\cite{zhong2024empowering} to incorporate a broader range of goal conditions, as the generalization capabilities of our framework critically depend on trajectory diversity. Our dataset $\mathcal{D}$ consists of trajectories $\mathcal{T}_t = (\mathcal{S}_t, \mathcal{O}_t, a_t, r_t, \mathcal{O}_{t+1}, \mathcal{S}_{t+1}, G_t^{final})$, where $t$ is the time step, $\mathcal{S}_t$ and $\mathcal{O}_t$ represent the tracker's state and observation, $a_t$ and $r_t$ denote the action and IoU-based reward, and $G_t^{final}$ specifies the spatial goal. For each episode, we randomly sample goals within the tracker's field of view to ensure diverse spatial configurations, with relative distances $\rho_t \in (200, \rho_{\text{max}})$ and angles $\theta_t \in (-\theta_{\text{max}}/2, \theta_{\text{max}}/2)$. A state-based PID controller with injected noise perturbations generates the goal-conditioned tracking trajectories, enhancing variability and robustness. The final training dataset comprises 10 million steps. 

\item \underline{IoU-based Training Reward.}
The reward function $r_t$ is designed to guide the policy move toward aligning $\mathcal{O}_t$ with $G_t^{final}$. At each time step $t$, the reward is defined as:
\begin{equation}
r_t = \text{IoU}(G_t^{final}, \mathcal{B}(\mathcal{VFM}(\mathcal{O}_t,\mathcal{C}_t))),
\label{eq:reward_function}
\end{equation}
where higher values of Intersection over Union (IoU) indicate the agent is moving towards better alignment in the 2D image. 

\item \underline{Offline Reinforcement Learning.}
In this paper, we extend from the standard offline RL algorithms, Conservative Q-Learning (CQL)~\cite{kumar2020conservative}, adapting to goal-conditioned setting. Specificallu, we use two critic networks $Q^{1}_\theta$, $Q^{2}_\theta$ to estimate goal-conditioned Q values.
The Q-functions are updated by minimizing the following objective:
$L_\theta = \sum^\mathbf{N}_{\mathbf{\mathcal{G}=1}} L_{\mathcal{G}}$, where the objective consists of the sum of the losses from all sampled goals in the batch, with each goal's loss computed according to the following formula:
%

\begin{align} 
L_{\mathcal{G}}
&=\mathbb{E}_{s}\left[\log \sum_{a} \exp Q^i_\theta(s, a)
-\mathbb{E}_{a \sim \pi_\phi(a \mid s)}[Q^i_\theta(s, a)]\right] \nonumber \\
&+ \frac{1}{2} \mathbb{E}_{s, a, s^{\prime}}\left[\left(Q^i_\theta(s, a) - \left( r+\gamma \mathbb{E}_{a^{\prime} \sim \pi_\phi}\left[Q_{min}(s^{\prime}, a^{\prime}) -\alpha \log \pi_\phi(a^{\prime} \mid s^{\prime})\right]\right) \right)^{2}\right]
\end{align} 
where $\theta$ and ${\phi}$ are network parameters, $\alpha$ is the entropy regularization coefficient which controls the degree of exploration. $i \in \{1,2\}$, $Q_{min}=min_{i \in \{1,2\}}Q_\theta^i$, $\gamma$ is the discount factor, $\pi_\phi$ is the learned policy that derived from the actor network $V_\phi$. 
Note that the state-goal aligner and the recurrent policy are jointly optimized by the RL loss. 

\item \underline{Auxiliary Reward Regression.}
For the goal-state aligner, we introduce an auxiliary reward regression task during training to facilitate the alignment between the goal and state representations. This task encourages the agent to recognize high-reward states associated with different goals, effectively steering the learned action policy toward these states. Specifically, we incorporate a fully connected layer following the goal-state aligner, predicting the alignment reward $\hat{r}_{t}$, the alignment loss is computed using $L_{\text{reg}} = \text{MSE}(r_t, \hat{r}_t),
\label{eq:reward_regression_loss}$
where $\text{MSE}$ denotes the mean squared error, the gradients are only updated to the goal-state aligner. Note that, the output of the fully connected layer will not be fed to the recurrent policy network, which will not affect inference stage.
}
\end{itemize}

\vspace{-0.3cm}
\section{Experiment}
\vspace{-0.3cm}
\label{exp}
In this section, we conduct comprehensive experiments in virtual environments and real-world scenarios, aiming to address the following \textbf{five} questions: 
RQ1) Can HIEVT outperform state-of-the-art models on tracking performance, robustness, and generalizability?
RQ2) How does the adjusting precision and efficiency of HIEVT when there is a new instruction?
RQ3) How do HIEVT and baselines perform under different target moving speeds?
RQ4) How do key components of HIEVT affect its performance?
RQ5) How does HIEVT perform in real-world environments?

\begin{figure}
    \centering
    \includegraphics[width=0.95\linewidth]{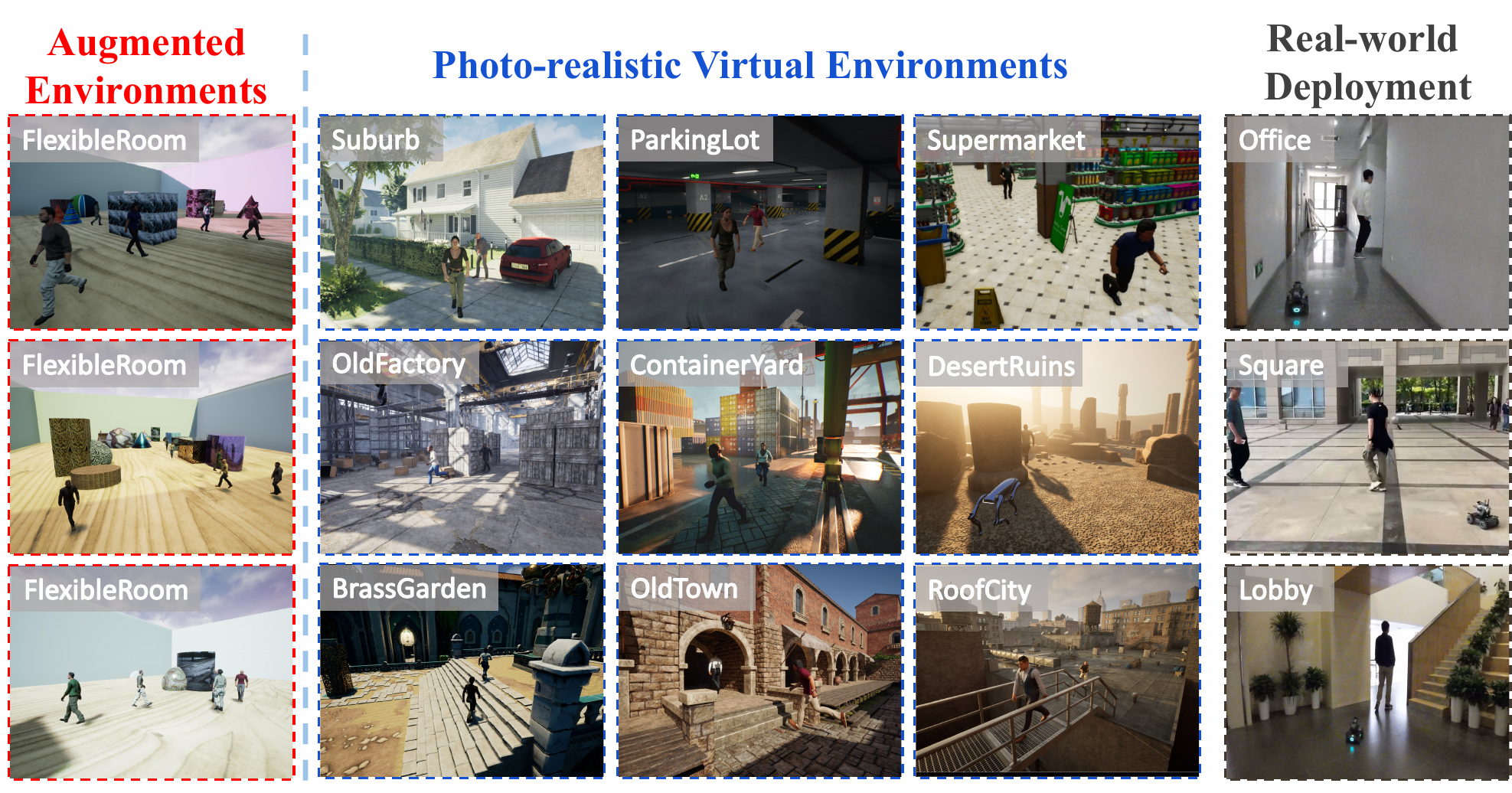}
    \vspace{-0.2cm}
    \caption{
    The examples of virtual and real-world environments used in our experiments. The \underline{FlexibleRoom} environment is used for training data collection, featuring diverse augmentable factor. the nine photo-realistic environments in the middle are used for quantitative evaluation, we also deploy our proposed method on three real-world scenarios to validate the effectiveness and transferability.}
    \label{fig:env}
    \vspace{-0.3cm}
\end{figure}

\subsection{Experimental Setup}

\paragraph{Environments}
We evaluate our approach across 10 virtual environments based on UnrealCV~\cite{qiu2017unrealcv}. \underline{FlexibleRoom}~\cite{zhong2024empowering} serves as our training environment, where the built-in navigation system generates diverse trajectories essential for goal-conditioned RL. For testing, we use \underline{FlexibleRoom} and other 9 photo-realistic environments that present complementary challenges: urban settings (\underline{Suburb}, \underline{Old Town}, \underline{Roof City}), confined spaces (\underline{SuperMarket}, \underline{Parking Lot}), industrial areas (\underline{Old Factory}, \underline{Container Yard}), and architectural complexes (\underline{Desert Ruins}, \underline{Brass Gardens}). These environments feature varied lighting conditions, occlusions, multi-level structures, and both indoor/outdoor transitions. Figure~\ref{fig:env} shows snapshots of these environments, with additional details in supplementary materials. In all environments, the tracker's visibility is defined by a 90-degree fan-shaped sector with a 750 cm radius ($\rho_{max}=750$, $\theta_{max}=90^\circ$).

\paragraph{Instruction Set Creation}
Due to computational constraints, we evaluate our system using a diverse but finite set of natural language instructions mapped to four representative spatial positions (detailed in supplementary materials. Our instruction set (listed in supplementary materials) includes both absolute spatial directives ("move to the left side") and relative positional commands ("move closer but stay to the right"). While these instructions contain inherent ambiguity, we ensure objective evaluation by calculating rewards based on pre-defined spatial goals corresponding to each instruction, allowing quantitative performance assessment despite the instructions' natural language variability.

\paragraph{Evaluation Metric}
Our ultimate goal is to realise the flexibility and versatility of embodied visual tracking and enable more natural human-robot interactions through our hierarchical aligned agents. Therefore, we keep the consistency with the task definition, assessing our method’s ability to track given diverse textual instructions.
In the experiment, we call the UnrealCV API to obtain the tracker's real-time ground truth spatial position $(\rho_t,\theta_t)$ and calculate the reward corresponding to different instructions, formulated as:
$r(I_t, s_t) = 1- \frac{\vert\rho_t - \rho_t^*\vert}{\rho_{max}} - \frac{\vert\theta_t-\theta_t^*\vert}{\theta_{max}}$, where $(\rho_{max}, \theta_{max})$ are the maximum distance and angle within the tracker's field of view, $(\rho_t^*, \theta_t^*)$ are spatial goal corresponding to Instruction $I_t$.
We follow the evaluation setting in previous works ~\cite{zhong2023rspt, zhong2024empowering}, setting the episode length to 500 steps with corresponding termination conditions. We use three evaluation metrics: 1) \underline{Average Accumulated Reward (AR)} calculates the average accumulated reward over 50 episodes, indicating overall performance in instruction-behavior alignment; 2) \underline{Average Episode Length (EL)} is the average number of steps across 50 episodes, reflecting long-term tracking performance. 3) \underline{Success Rate (SR)} calculates the percentage of episodes reaching 500 steps in 50 episodes. 

\paragraph{Baselines}
To comprehensively evaluate the tracking performance of our proposed method under instruction  settings, we compare it with four representative baseline methods:
\noindent\textbf{1) Bbox-based PID:} Uses rule-based bounding box selection according to instructions, with a PID controller maximizing IoU between the target mask and the selected spatial goal, serving as a straightforward benchmark. \noindent\textbf{2) Ensembled RL:} Extends the state-of-the-art EVT model~\cite{zhong2024empowering} by training multiple policies under different goal settings $(\rho^*,\theta^*)$. During evaluation, we select the policy that best matches the inferred goal from the instruction. \noindent\textbf{3) GPT-4o:} Directly generates discrete actions based on the observed image and instruction, leveraging multimodal reasoning to interpret visual and textual information jointly. \noindent\textbf{4) OpenVLA:} We fine-tune OpenVLA~\cite{kim2024openvla} on our tracking dataset, using text instructions as goal representations to generate actions from image observations. Unlike navigation approaches that rely on step-by-step reasoning with large models (unsuitable for real-time tracking), OpenVLA attempts to balance prior knowledge with real-time performance requirements. Implementation details for all baselines are available in supplementary materials.

\begin{table}
    \centering
    \caption{Performance comparison across ten environments with dynamic instruction changes. Each cell reports Average Accumulated Reward (AR), Average Episode Length (EL), and Success Rate (SR) in the format AR/EL/SR. During each episode, the user instruction randomly changes four times, testing adaptation to shifting goals. Our method maintains high performance (SR $\geq 0.58$) across all environments while matching the real-time performance (50 FPS) of traditional methods.}

    \label{tab:main}
     \resizebox{\linewidth}{!}{
    \begin{threeparttable}
    \begin{tabular}{l|c|ccccc}
    \hline
         \multirow{2}{*}{Method} & \multirow{2}{*}{FPS} & \multicolumn{5}{c}{Environment Name} \\
         \cline{3-7}
         & & \textbf{FlexibleRoom} & Suburb & Supermarket & Parking Lot & Old Factory \\
    \hline
        Bbox-based PID & 50 & 154/365/0.5 & 124/386/0.36 & 175/298/0.26 & 112/356/0.38 & 80/309/0.3 \\
        Ensembled RL  & 50 & 183/330/0.36 & 126/294/0.22 & 114/393/0.38 & 169/301/0.42 & 64/334/0.36 \\
        OpenVLA & 8 & 123/253/0.25 & 63/311/0.18 & 68/233/0.1 & 12/205/0.08 & 5/226/0.12 \\
        GPT-4o & 0.38 & 15/194/0.1 & 14/241/0.20 & 54/286/0.16 & 23/286/0.3 & 5/297/0.32 \\
        Ours & 50 & 278/500/1.0 & 166/445/0.72 & 212/422/0.64 & 169/491/0.93 & 128/469/0.84 \\
    \hline
         \multirow{2}{*}{Method} & \multirow{2}{*}{FPS} & \multicolumn{5}{c}{Environment Name} \\
         \cline{3-7}
         & & Container Yard & Desert Ruins & Brass Garden & Modular Old Town & Roof City \\
    \hline
        Bbox-based PID & 50 & 140/369/0.42 & 128/297/0.26 & 120/302/0.38 & 73/305/0.32 & 64/289/0.34 \\
        Ensembled RL & 50 & 27/327/0.24 & 56/368/0.34 & -8/348/0.34 & 14/356/0.32 & 19/322/0.38 \\
        OpenVLA & 8 & 43/232/0.08 & 22/219/0.12 & 30/236/0.08 & -9/213/0.16 & 96/303/0.22 \\
        GPT-4o & 0.38 & -43/149/0.0 & -5/249/0.26 & 3/243/0.18 & 29/304/0.28 & 18/256/0.2 \\
        Ours & 50 & 156/469/0.76 & 148/434/0.74 & 126/425/0.72 & 85/433/0.64 & 86/400/0.58 \\
    \hline
    
    \end{tabular}
    \end{threeparttable}}

\end{table}

\begin{figure}[!t]
    \centering
    \begin{minipage}[t]{0.48\linewidth}
        \centering
            \caption{ Pixel-level distance between target center and goal center across time steps. The spike at step \#101 represents a goal shift instruction, followed by rapid corrections (steps \#107, \#112) as our agent adjusts to the new spatial goal.}
        \includegraphics[width=\linewidth]{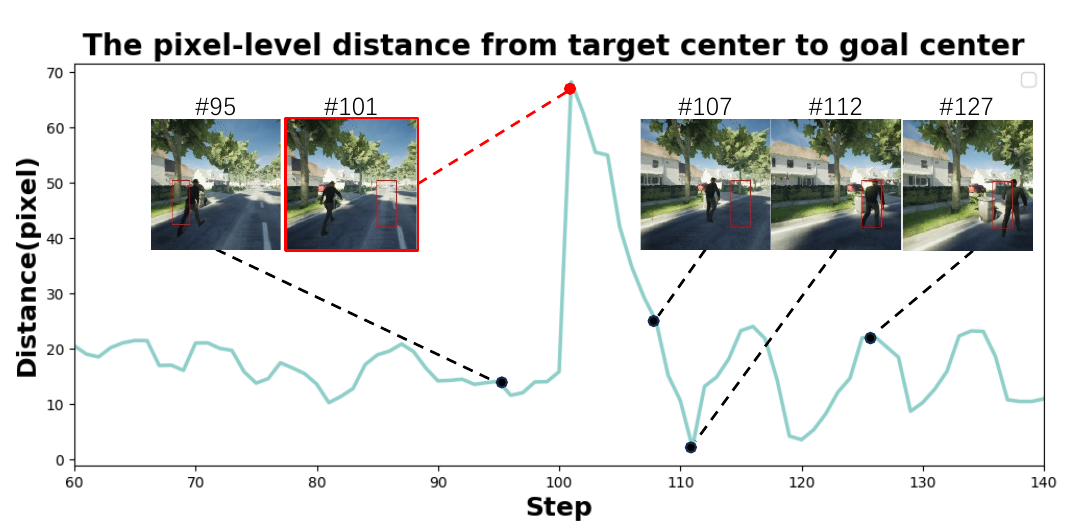}
    
        \label{fig:tracker_output}
    \end{minipage}%
    \hfill
    \begin{minipage}[t]{0.5\linewidth}
        \centering
        \vspace{0pt}  
        \captionof{table}{\textbf{Speed robustness analysis} in FlexibleRoom. Values: AR/EL/SR. Only our method maintains high success rates at high speeds (SR=0.84 at 2.0 m/s) while baselines show severe degradation.}
        \label{tab:speed_comparison}
        \vspace{2mm}
        \scriptsize
        \setlength{\tabcolsep}{3pt}
        \begin{tabular}{l|ccc}
        \toprule
        \multirow{2}{*}{\textbf{Method}} & \multicolumn{3}{c}{\textbf{Target Speed}} \\
        \cmidrule{2-4}
         & 0.5 m/s & 1.0 m/s & 2.0 m/s \\ 
        \midrule
        Bbox-based PID & 154/365/0.50 & -5/330/0.44 & -23/125/0.02 \\
        Ensembled Policy & 183/330/0.36 & 187/313/0.32 & -183/208/0.10 \\
        OpenVLA & 123/253/0.25 & 117/249/0.15 & -34/96/0.00 \\
        GPT-4o & 15/194/0.10 & 4/229/0.08 & -119/122/0.00 \\
        
        \textbf{Ours} & \textbf{278/500/1.00} & \textbf{274/496/0.98} & \textbf{145/463/0.84} \\ 
        \bottomrule
        \end{tabular}
    \end{minipage}
    \vspace{-5mm}
\end{figure}

\subsection{Main Results~(RQ1)}

\paragraph{Overall Performance}
Our method consistently outperforms all baselines by significant margins, as shown in Table ~\ref{tab:main}. In the \underline{FlexibleRoom} training environment, we achieve a perfect success rate (1.0) with an average reward of 278, compared to the next best baseline (Ensembled RL) at 183. While Box-based PID, Ensembled RL, and OpenVLA demonstrate basic instruction-aware tracking (EL = 365, 330, and 253 respectively), their substantially lower AR and SR metrics reveal significant limitations in dynamic alignment and long-term tracking capabilities.

\paragraph{Generalization Across Environments}
More importantly, our approach maintains robust performance across all nine unseen environments, with success rates between 0.58 (\underline{Roof City}) and 0.93 (\underline{Parking Lot}) and episode lengths consistently exceeding 400 steps. This environmental-agnostic performance contrasts sharply with all baselines, which show severe degradation in unseen settings.

\paragraph{Baseline Limitations}
The comparative analysis reveals distinct failure modes across baselines: (1) Bbox-based PID suffers from parameter rigidity, causing accumulated errors during goal transitions (SR varies from 0.26 to 0.42); (2) Ensembled RL experiences temporal feature inconsistency during policy switching, resulting in goal-behavior misalignment despite reasonable environmental generalization; (3) OpenVLA lacks explicit spatial goal representations, severely limiting cross-domain generalization; and (4) GPT-4o, despite strong reasoning capabilities, fails due to extreme inference latency (0.38 FPS), making it unsuitable for real-time tracking—a limitation further confirmed in our speed comparison experiments (Table~\ref{tab:speed_comparison}).

\paragraph{Key Advantages}
Our approach's superior performance stems from its spatial goal representation serving as an intermediate abstraction between language and action. This design enables consistent tracking across diverse environments without requiring environment-specific adaptation, while maintaining real-time performance (50 FPS). The results demonstrate that our framework successfully bridges the gap between natural language instructions and precise spatial tracking behaviors in dynamic, real-world scenarios.

\subsection{Adaptability of the goal-aligned Policy ~(RQ2)}
\paragraph{Precise and Fast Alignment} To demonstrate HIAEVT's precise and efficient adaptation to dynamically changed instructions, we visualised a goal-switch case and recorded the pixel-level deviation for quality analysis, as shown in Figure ~\ref{fig:tracker_output}. When the goal changes at step \#101, our system responds with remarkable speed—reducing the pixel distance from 67 to 24 pixels within just 6 steps (\#107) and achieving near-perfect alignment (2 pixels) by step \#112. This represents complete adaptation within \textbf{220ms} of real-time operation. The system maintains this precision through step \#127 despite continued target movement, showcasing both initial responsiveness and sustained tracking accuracy. Table ~\ref{tab:speed_comparison} further validates this capability across varying target speeds, with our method maintaining a 0.84 success rate even at 2.0 m/s while all baselines fail (0.00-0.10 success rates). This performance stems from our mask-based goal representation enabling the policy to rapidly interpret and align with new spatial objectives—a fundamental advantage over traditional approaches that struggle with dynamic instruction scenarios.
\paragraph{Real-time inference challenges ~(RQ3)}
We tested system robustness by increasing target speeds from $0.5m/s$ to $2.0m/s$ (Table~\ref{tab:speed_comparison}). This experiment reveals a critical real-world challenge: making decisions under real-time constraints. Large-model based methods (OpenVLA, GPT-4o) fail completely at higher speeds (0.00 SR at 2.0m/s) due to inference latency (8 FPS, 0.38 FPS), while conventional approaches (PID, Ensembled Policy) degrade significantly (0.50→0.02 SR, 0.36→0.10 SR). In contrast, our method maintains a high success rate (0.84 SR) even at 2.0m/s. This demonstrates our hierarchical framework successfully balances semantic understanding with operational efficiency—a crucial capability for real-world deployment where targets move at unpredictable speeds and instructions require immediate responses.

\vspace{-0.2cm}
\subsection{Ablation Studies ~(RQ4)}
We conduct ablation studies (Table~\ref{tab:ablation}) to verify each component's contribution. 
Our findings reveal: 1) \textbf{Goal representation}: Vector-based goals maintain reasonable episode lengths but limit precise spatial strategy learning. CLIP-encoded text goals perform poorly due to CLIP's limited semantic-visual aligning capabilities, highlighting the superiority of our approach. 2)\textbf{Training objectives}: Removing reward regression or IoU-based rewards significantly decreases performance during goal transitions, demonstrating their critical role in maintaining dynamic adaptability. 3) \textbf{LLM scaling}: While larger models (GPT-4o vs. Gemma2-27B) generally provide better instruction-goal alignment, even smaller models (Gemma2-2B) achieve strong results, indicating our framework's efficiency across computational constraints.

\vspace{-0.3cm}
\subsection{Results on Real-world Environments ~(RQ5)}
\begin{figure*}[t]
    \centering
    \includegraphics[width=0.9\linewidth]{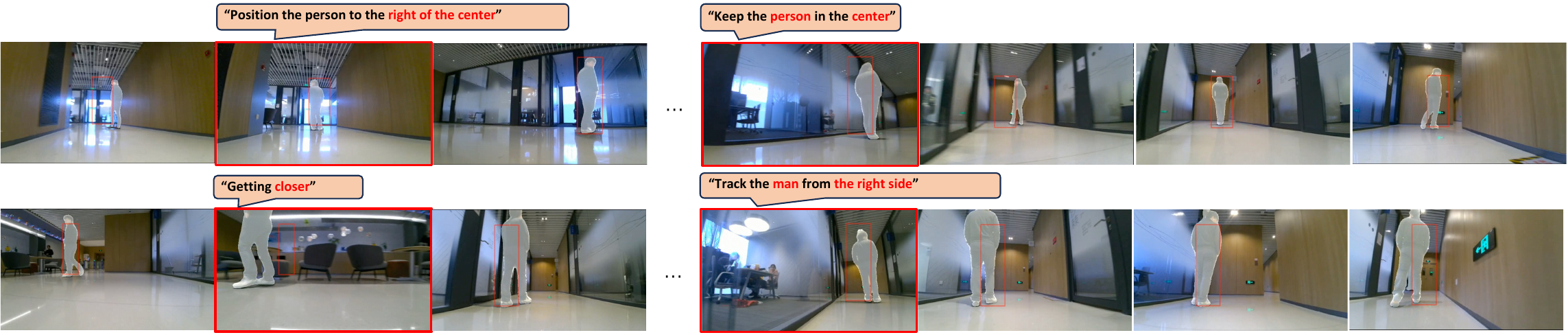}
    \caption{We deploy the agent into a wheel robot in a real-world scenario. The sequence shows the robot responding to the user's textual instruction with real-time adjustments in tracking and positioning. Visual annotation (white color mask) is generated by SAM-Track with text prompt ``person'', the red bounding box is generated by LLM parser based on instructions and observation.}
    \label{fig:real-world}
    \vspace{-0.3cm}
\end{figure*}

To demonstrate the practical applicability and robustness of our goal-behavior alignment framework, we deploy the agent on a mobile wheel robot to handle real-world variability and dynamically adapt to human instructions, the hardware information and more video clips are available in supplementary materials. We use SAM-Track~\cite{cheng2023segment} to generate segmentation masks from the robot's real-time observations. As shown in Figure ~\ref{fig:real-world}, we test our agent in a daily work area, assigning different text instructions during tracking. Each row presents two different textual instructions, with our model generating reasonable bounding box positions and dynamically adjusting the target to the desired location. In the second row, when the target made a sudden turn at the beginning, our model demonstrated real-time responsiveness, quickly adjusting its position to maintain tracking and successfully executing the user’s instructions afterward.

\begin{table}[!t]
\centering
\caption{Ablation study and model variant performance comparison with different numbers of instruction switches per episode in FlexibleRoom environment. Each cell shows average reward/episode length/success rate.}
\label{tab:ablation}
\begin{tabular}{l|cccc}
\hline
\multirow{2}{*}{Method} & \multicolumn{4}{c}{Number of Instruction Switches per Episode} \\
\cline{2-5}
 & 1 & 2 & 3 & 4 \\
\hline
w/o spatial goal(Vector) & 100/497/0.48 & 44/470/0.44 & 77/477/0.46 & 27/487/0.46 \\
 w/o spatial goal (CLIP) & 156/307/0.36 & 124/280/0.26 & 131/302/0.26 & 102/244/0.24 \\
 w/o reward regression & 4/465/0.82 & -4/402/0.56 & 6/398/0.58 & 16/378/0.46 \\
 w/o IOU-based reward & 109/426/0.78 & 66/393/0.68 & -10/298/0.36 & -1/341/0.42 \\
\hline
\hline
Ours (Gemma2-2B) & 240/456/0.84 & 183/441/0.80 & 133/421/0.70 & 182/459/0.84 \\
Ours (Gemma2-27B) & 243/467/0.78 & 216/455/0.80 & 116/415/0.64 & 184/465/0.80 \\
Ours (GPT-4o)& \textbf{278}/ 492/ \textbf{0.94} & \textbf{274}/ \textbf{489}/ \textbf{0.92} &\textbf{275}/ \textbf{488}/ \textbf{0.96} &\textbf{274}/ \textbf{496}/ \textbf{0.98} \\ 
\hline
\end{tabular}
\vspace{-0.3cm}
\end{table}

\vspace{-0.3cm}
\section{Conclusion}
\vspace{-0.2cm}
\label{conclusion}

In this paper, we introduced HIEVT, a hierarchical tracking agent for User-Centric Embodied Visual Tracking that bridges the semantic-spatial gap in human-robot interaction. By decoupling language understanding from motion control through intermediate  spatial goals, our approach combines the reasoning capabilities of large language models with the real-time performance demands of dynamic tracking. Experimental results across ten diverse environments demonstrate substantial performance advantages over existing methods, particularly in adaptability and generalization to unseen environments. Our successful real-world robot deployment validates that this hierarchical design effectively balances sophisticated instruction understanding with operational efficiency, providing an elegant and practical solution for user-guided spatial intelligence in embodied systems that scales with minimal additional data requirements.

\small

\bibliographystyle{unsrt}
\bibliography{ref}

\medskip

\clearpage

\appendix
\section{Related Works} \label{sec:related}

\textbf{Embodied Visual Tracking (EVT)} is a foundational skill of embodied AI. It has been a focal point for improving agents' robustness and generalization in complex environments, including unseen targets~\cite{zhong2018advat,zhong2019ad}, distractions~\cite{zhong2021distractor}, and unstructured maps~\cite{zhong2023rspt}. Recent work~\cite{zhong2024empowering} combines the visual foundation model~\cite{cheng2023tracking} and offline reinforcement learning to improve the training efficiency and generalization of the tracker. However, these methods train the agent to track at a specific relative position to the target. If we need to change the goal, fine-tuning the policy network is required to adapt to new goals. This limits the flexibility and applicability of the agents across varied scenarios and tasks. 
In contrast, we take into account the online interaction between the user and the agent. We build an instruction-aware agent with an LLM-based instruction parser and an RL-based goal-oriented policy to efficiently adapt to various user instructions. As a result, our agent offers more flexible and customized tracking capabilities to enhance its responsiveness to a wide range of demands.

\noindent
\textbf{Instruction-aware Robot} is developed to complete tasks by understanding and following human instructions. The motivation behind human instruction to action lies in the need to bridge the gap between high-level human intentions and low-level robotic actions, thereby enhancing human-robot collaboration. Recent advancements in natural language processing and computer vision have garnered significant attention from the embodied AI community. 
Touchdown~\cite{chen2019touchdown} and R2R ~\cite{anderson2018vision} are early works that use human instructions to navigate in unseen real-world environments. IVLN ~\cite{krantz2023iterative} integrates with past memory and explicit maps to enhance vision-language navigation ability. Similarly, PALM-E ~\cite{driess2023palm} builds on a large language model, fusing language instructions into reasoning and adapting it to various embodied tasks. 
Then AVLEN ~\cite{paul2022avlen} first attempted to use audio and natural language to improve embodied navigation in 3D environments. In addition, in robotic manipulation, researchers have further integrated large models and natural, multimodal instructions to enhance capabilities to understand and execute complex tasks. For example, VIMA ~\cite{jiang2022vima} integrates multi-modal prompts for systematic generalization, PALM-E ~\cite{driess2023palm} and OpenVLA~\cite{kim2024openvla} build on a large pre-trained model and adapt it to various embodied tasks.  
However, the direct application of these models to embodied visual tracking presents challenges, given the need for an agent capable of rapidly adjusting its behavior in response to user demands in dynamic environments.

\section{Virtual Environment}
\label{app:env}

Our experiments were conducted across 10 diverse virtual environments built using Unreal Engine and integrating UnrealCV ~\cite{qiu2017unrealcv} for programmatic control. These environments were developed based on UnrealZoo ~\cite{zhong2024unrealzooenrichingphotorealisticvirtual}, aiming to evaluate different aspects of instruction-aware tracking under various challenging conditions. Environment binary could be downloaded from \url{https://modelscope.cn/datasets/UnrealZoo/UnrealZoo-UE4}, and code are available in \url{https://anonymous.4open.science/r/Instruction_Aware_Tracking-31B8/}
\subsection{Training Environment}
FlexibleRoom: This environment, adopted from previous work ~\cite{zhong2023rspt,zhong2021distractor}, serves as our primary training venue. It features an adaptable indoor space with programmable lighting conditions, furniture layouts, and target navigation patterns. The environment's built-in navigation system enables automatic generation of diverse trajectories through randomly sampled destinations, which is particularly valuable for data collection in goal-conditioned reinforcement learning. We extended this environment with customizable appearance factors like texture, lighting, and furniture placement to enhance training diversity and reduce overfitting. The modular design allows us to systematically control visual complexity while maintaining consistency in the underlying spatial relationships.
\subsection{Testing Environments}
\textbf{Suburb:} A meticulously designed suburban neighborhood featuring irregular terrain, diverse vegetation, and dynamic obstacles that simulate pedestrian and vehicular movement. This environment tests the agent's ability to maintain tracking across changing elevation, lighting conditions, and partial occlusions from trees and structures. The open spaces combined with clustered obstacles create complex tracking scenarios with variable target visibility.

\textbf{Supermarket:} An indoor retail environment with intricate item shelves, static displays, and narrow aisles that closely mimic real-world shopping scenarios. The dense arrangement of objects creates numerous occlusion challenges and confined spaces for navigation. This environment evaluates the system's performance in crowded indoor settings where the target frequently disappears behind shelves and reappears elsewhere.

\textbf{Parking Lot:} An outdoor environment featuring multiple parked vehicles under dim lighting conditions. The uniform structure combined with low visibility areas and complex shadows tests the agent's ability to discriminate targets in visually challenging scenes. The environment transitions between open areas and confined spaces between vehicles, requiring adaptive tracking strategies.

\textbf{Old Factory:} A deteriorated industrial setting characterized by numerous steel pillars, scattered wooden crates, and uneven lighting. The environment features high ceilings with exposed structural elements and complex shadows that create challenging visual conditions. The combination of open factory floor areas and cluttered storage zones tests the agent's ability to track across rapidly changing visual contexts.

\textbf{Container Yard:} A dynamic logistics environment featuring stacked shipping containers under changing lighting conditions. The geometric regularity of the containers combined with dramatic lighting variations creates challenging perception scenarios. The environment features narrow corridors between container stacks that frequently occlude targets, testing the system's ability to predict movement through temporary visual obstruction.

\textbf{Desert Ruins:} An archaeological site set in harsh desert lighting conditions with scattered walls and pillars creating a complex spatial layout. The environment combines open areas with confined passages and features extreme lighting contrasts between shadow and direct sunlight. This tests the agent's robustness to challenging lighting conditions and irregular spatial structures.

\textbf{Brass Gardens:} A palace-style architectural complex featuring narrow corridors and multi-level platforms connected by staircases. This environment uniquely tests non-planar tracking capabilities, as targets frequently change elevation while moving through the environment. The ornate architectural elements and varying ceiling heights create complex spatial reasoning challenges for maintaining consistent tracking.

\textbf{Modular Old Town:} A European-style hillside village with interconnected indoor and outdoor spaces linked by narrow, undulating stairways. This environment combines both open plazas and confined interior spaces, requiring frequent adaptation to changing spatial contexts. The irregular layout with multiple elevation changes tests the agent's ability to maintain tracking continuity across diverse architectural spaces.

\textbf{Roof City:} A rooftop cityscape featuring protruding air ducts, walkways, and scattered debris that create a maze-like environment. The constrained navigation paths combined with varying elevation levels test the agent's ability to predict movement in spatially restricted areas. The urban setting also introduces complex background textures and challenging lighting conditions from reflective surfaces.

\subsection{Goal Randomization}
\label{app:goal_randomization}
During training, we implement goal randomization within a continuous space, within the range of $\rho^*\in(200,600)$ and $\theta^*\in(-25^\circ,25^\circ)$. However, due to computational constraints, it is impractical to evaluate all points within this continuous space. Therefore, for evaluation, we manually select four representative discrete goal points as a goal list: $[g_{close}, g_{far}, g_{left}, g_{right}]$. In the experimental section, each random goal switch is performed by sampling without replacement from this goal list. Four discrete goal are as follows: $g_{close}=[200,0^\circ]$, representing a scenario where the target should remain close to the tracker and centered in its field of view; $g_{far}=[450,0^\circ]$, requiring the target to be farther away while still centered; $g_{left}=[350,-20^\circ]$,  where the target is positioned to the left; $g_{right}=[350,20^\circ]$, where the target is positioned to the right.
\subsection{Instructions}
\label{app:instruction_list}
To fairly and accurately evaluate our method's ability to align with high-level text instructions, we generated a list of text instructions representing various spatial objectives, as shown in Table ~\ref{tab:instructions}. These instructions set are extend from previous mentioned four discrete goal positions, balancing natural language variation with consistent evaluation metrics.These instructions are designed to reflect both absolute positions and relative position changes, including:
\begin{itemize}
    \item \textbf{Absolute spatial directives:} Instructions specifying explicit positions, from line 0 to line 22 in Table ~\ref{tab:instructions}.
    \item \textbf{Relative positional adjustments:} Instructions specifying changes relative to current position, from line 22 to line 42 in Table ~\ref{tab:instructions}.

\end{itemize}

The core idea is that if the model correctly interprets the text instruction, the agent's step reward $r(g_t^\prime, s_t)$ should closely match the reward value $r(g_t^*, s_t)$ associated with the intended spatial goal.
Our approach emphasizes understanding vague and abstract human intentions rather than specifying exact spatial locations. Therefore, a small margin of deviation between the reward generated by following the text instruction and the reward for the exact goal position $r(g_t^*, s_t)$ is allowed. This method helps us assess how well the model captures the intended meaning of the instructions and translates them into appropriate actions.

\begin{table*}[!t]
\centering
\caption{Examples of Textual Instructions with Corresponding Spatial Goals}
\label{tab:instructions}
\begin{tabular}{c|c}
\hline
 \textbf{Example Instruction} & \textbf{Corresponding Spatial Goal} \\ \hline
 ``Keep the person in the close center.'' & $(\rho^*, \theta^*) = (200, 0^\circ)$ \\ \hline
 ``Ensure the person stays at the close center.'' & $(\rho^*, \theta^*) = (200, 0^\circ)$ \\ \hline
``Maintain the person in the center at close range.'' & $(\rho^*, \theta^*) = (200, 0^\circ)$ \\ \hline 
``Keep the person positioned near the center.'' & $(\rho^*, \theta^*) = (200, 0^\circ)$ \\ \hline 
``Ensure the person remains close to the center.'' & $(\rho^*, \theta^*) = (200, 0^\circ)$ \\ \hline

 ``Keep the person in the far-away center.'' & $(\rho^*, \theta^*) = (450, 0^\circ)$ \\ \hline
 ``Ensure the person stays at the far-away center.'' & $(\rho^*, \theta^*) = (450, 0^\circ)$ \\ \hline
``Keep the person in the center and at a far distance.'' & $(\rho^*, \theta^*) = (450, 0^\circ)$ \\ \hline 
``Keep the person in the far center.'' & $(\rho^*, \theta^*) = (450, 0^\circ)$ \\ \hline
``Ensure the person stays at the far central position.'' & $(\rho^*, \theta^*) = (450, 0^\circ)$ \\ \hline

 ``Keep the person on the left.'' & $(\rho^*, \theta^*) = (350, -20^\circ)$ \\ \hline
 ``Position the person to the left of the center.'' & $(\rho^*, \theta^*) = (350, -20^\circ)$ \\ \hline 
 ``Keep the person to the left of the center.'' & $(\rho^*, \theta^*) = (350, -20^\circ)$ \\ \hline 
  ``Ensure the person remains on the left side.'' & $(\rho^*, \theta^*) = (350, -20^\circ)$ \\ \hline  ``Position the person on the left side of the center.'' & $(\rho^*, \theta^*) = (350, -20^\circ)$ \\ \hline  
  ``Keep the person aligned to the left.'' & $(\rho^*, \theta^*) = (350, -20^\circ)$ \\ \hline

 ``Keep the person on the right.'' & $(\rho^*, \theta^*) = (350, 20^\circ)$ \\ \hline
 ``Position the person to the right of the center.'' & $(\rho^*, \theta^*) = (350, 20^\circ)$ \\ \hline 
 ``Keep the person to the right of the center.'' & $(\rho^*, \theta^*) = (350, 20^\circ)$ \\ \hline 
  ``Ensure the person remains on the right side.'' & $(\rho^*, \theta^*) = (350, 20^\circ)$ \\ \hline  ``Position the person on the right side of the center.'' & $(\rho^*, \theta^*) = (350, 20^\circ)$ \\ \hline  
  ``Keep the person aligned to the right.'' & $(\rho^*, \theta^*) = (350, 20^\circ)$ \\ \hline

 ``Keep the person in the current direction but closer.'' & $(\Delta \rho, \Delta \theta) = (-150, 0^\circ)$ \\ \hline
 ``Move the person closer.'' & $(\Delta \rho, \Delta \theta) = (-150, 0^\circ)$ \\ \hline 
 ``Move the person closer while maintaining the same direction.'' & $(\Delta \rho, \Delta \theta) = (-150, 0^\circ)$ \\ \hline 
``Keep the person in the same direction, but reduce the distance.'' & $(\Delta \rho, \Delta \theta) = (-150, 0^\circ)$ \\ \hline 
``Bring the person closer, keeping the same direction.'' & $(\Delta \rho, \Delta \theta) = (-150, 0^\circ)$ \\ \hline

``Keep the person directly in front at a greater distance.'' & $(\Delta \rho, \Delta \theta) = (150, 0^\circ)$ \\ \hline
``Move further away.'' & $(\Delta \rho, \Delta \theta) = (150, 0^\circ)$ \\ \hline 
``Keep the person ahead at a greater distance.'' & $(\Delta \rho, \Delta \theta) = (150, 0^\circ)$ \\ \hline 
``Increase the distance between you and the front person.'' & $(\Delta \rho, \Delta \theta) = (150, 0^\circ)$ \\ \hline 
``Maintain the person in front, far from you.'' & $(\Delta \rho, \Delta \theta) = (150, 0^\circ)$ \\ \hline

 ``Track the person along the left side'' & $(\Delta \rho, \Delta \theta) = (0, -20^\circ)$ \\ \hline
 ``Shift a bit to the left.'' & $(\Delta \rho, \Delta \theta) = (0, -20^\circ)$ \\ \hline
 ``Move slightly to the left.'' & $(\Delta \rho, \Delta \theta) = (0, -20^\circ)$ \\ \hline
  ``Move a bit toward the left.'' & $(\Delta \rho, \Delta \theta) = (0, -20^\circ)$ \\ \hline
   ``Shift just a little to the left.'' & $(\Delta \rho, \Delta \theta) = (0, -20^\circ)$ \\ \hline

``Track the target along the right side.'' & $(\Delta \rho, \Delta \theta) = (0, 20^\circ)$ \\ \hline
 ``Shift a bit to the right.'' & $(\Delta \rho, \Delta \theta) = (0, 20^\circ)$ \\ \hline
``Move slightly to the right.'' & $(\Delta \rho, \Delta \theta) = (0, 20^\circ)$ \\ \hline
  ``Move a bit toward the right.'' & $(\Delta \rho, \Delta \theta) = (0, 20^\circ)$ \\ \hline
   ``Shift just a little to the right.'' & $(\Delta \rho, \Delta \theta) = (0, 20^\circ)$ \\ \hline

\end{tabular}
\end{table*}

\section{Data Collection}
\label{app:data_collection}
In our experiment setting, the tracking goal's relative spatial position is constrained within the range of $\rho^*\in(200,600)$ and $\theta^*\in(-25^\circ,25^\circ)$. We uniformly sample tracking goals from this range, resulting in an offline dataset of \textbf{1,750,000} steps used to train our proposed method. In this section, we introduce the details of the data collection process, including player initialization, and state-based PID controller with noise perturbation for tracker and trajectory generation. 

\subsection{Player Initialization}
\begin{figure}
    \centering
    \includegraphics[width=0.6\linewidth]{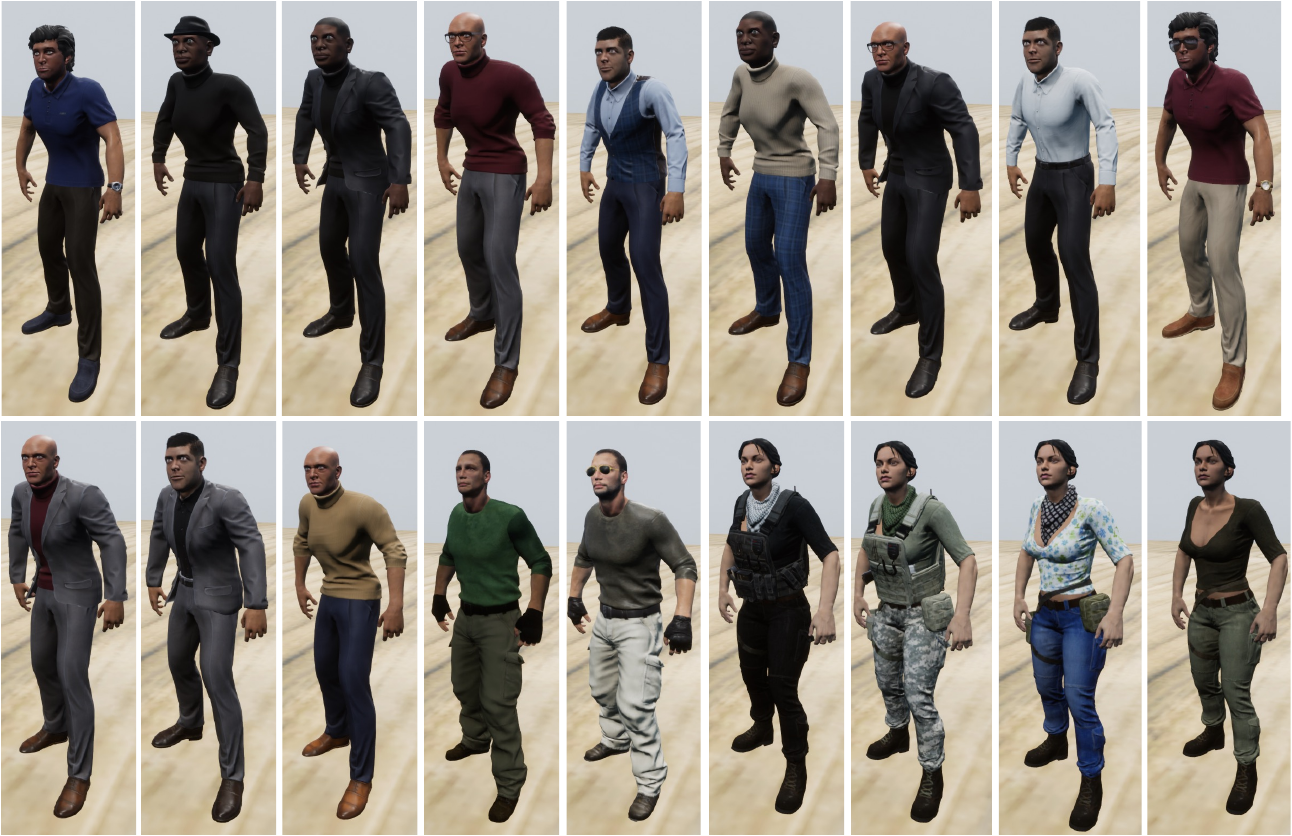}
    \caption{18 humanoid models are used in data collection and evaluation.}
    \label{fig:player}
\end{figure}

We use 18 humanoid models as targets and trackers, as shown in Figure~\ref{fig:player}. At the beginning of each episode, we randomly sample their appearance from the 18 humanoid models, and the target player will be randomly placed in the tracker's visible region.

\subsection{State-based PID Controller with Multi-level Perturbation}
\label{app:multi-level_pid}
We first use PID controllers to enable the agent to follow a target object, maintaining a specific distance and relative angle(e.g., 3 meters directly in front of the agent). The process is as follows:
\begin{itemize}
\item Setpoints: Define the desired distance and angle as the setpoints for the PID controller (e.g., 3 meters for distance and 0 degrees for angle).
\item Process Variables: Measure the actual distance and angle between the agent and the target object using the grounded state data accessible via the UnrealCV API. These measurements serve as the process variables for the PID controller.
\item Error Calculation: Calculate the error between the setpoints and the process variables, which will be used as inputs for the PID controller.
\item Control Output: Apply the PID equation to generate the control output, determining the agent's speed and direction:
    \begin{equation}
        u(t)=K_p e(t)+K_i \int_0^t e(\tau) d \tau+K_d \frac{d e(t)}{d t}
    \end{equation}
where $u(t)$ is the control output, $e(t)$ is the error, $K_p$, $K_i$, and $K_d$ are the proportional, integral, and derivative gains, respectively.
\item Fine-tune the PID gains to achieve optimal controller performance. For instance, increasing $K_p$ will enhance the agent's responsiveness to errors but may introduce overshoot or oscillations. Raising $K_i$ helps minimize steady-state error but can lead to integral windup or slower response. Boosting $K_d$ reduces overshoot and dampens oscillations but may also amplify noise or cause derivative kick. If the control output exceeds the defined action space limits, it is clipped. The tuned gains are detailed in Table ~\ref{tab:pid}.
\begin{table}[h]
        \centering
        \caption{The parameters we used in the state-based PID controller.}
        \begin{tabular}{cccc}
        \hline
        Controller & $K_p$ & $K_i$ & $K_d$ \\
        \hline
        Speed & 5 & 0.1 & 0.05\\
        Angle    & 1 & 0.01 & 0\\
        \hline
        \end{tabular}
        \label{tab:pid}
    \end{table}
\end{itemize}

Then, we introduce noise perturbation to the PID output, causing the agent to alternately deviate from and recover towards the desired distance and angle. This set-up aims to collect trajectories with diverse step rewards, alleviating the overestimation problem during offline training. We set a threshold $p=0.15$, and if the probability value at time $t$ is greater than $p$, which is $P(t)>p$, the agent takes a random action from the action space and continues for $L$ steps. We also adopt a random strategy to set the step length $L$, with an upper limit of 4. After the random actions of the agent end, we use a random function to determine the duration of the next random action $L$. Here, we set the upper limit to 4 because we found that the number of times the agent failed in a round significantly increased beyond 4. Therefore, we empirically set the upper limit of the random step length to 4.

\section{Implementation Details}
In this section, we detail the implementation of our basic setup, baseline methods, and proposed policy network structures.
\begin{table}[h]
        \centering
        \caption{The fine-tuned parameter for Bbox-based PID Controller.}
        \begin{tabular}{cccc}
        \hline
        Controller & $K_p$ & $K_i$ & $K_d$ \\
        \hline
        Speed & 0.2 & 0.01 & 0.03\\
        Angle    & 0.05 & 0.01 & 0.1\\
        \hline
        \end{tabular}
        \label{tab:bbox_pid}
\end{table}
\subsection{Basic Setting}
In our experiments, we utilize a continuous action space for agent movement control. The action space comprises two variables: the angular velocity, ranging from($-30^{\circ}/s$, $30^{\circ}/s$) and the linear velocity, ranging from ($-1\ m/s$ ,$1\ m/s$). We train our models using the Adam optimizer with a learning rate of 3e-5 and a batch size of 128.
\subsection{Baseline Methods}
\label{app:baselines}
\underline{Bbox-based PID:} We use the bounding box as the goal representation, adjusting the agent’s actions to maximize the intersection-over-union (IoU) with the target mask. We implemented two PID controllers to jointly control the agent's movement: one for linear velocity and one for angular velocity. For linear velocity($V_{linear}$), we use the areas of the target bounding box mask $A_{s}$ detected from state $s$, and goal bounding box masks $A_{goal}$ as input variables, the $V_{linear}$ is calculated as:
\begin{equation}
\begin{aligned} 
    V_{linear}(t)=
    &K_p (A_{goal}-A_{s}(t))+K_i \int_0^t (A_{goal}-A_{s}(\tau)) d \tau \\
    &+K_d \frac{d (A_{goal}-A_{s}(t))}{d t}
\end{aligned}
\end{equation}
where $K_p, K_i$, and $K_d$ are the proportional, integral, and derivative gains, respectively. The linear velocity is constrained within the range ($-100$, $100$) to match the training setup. A positive $V_{linear}$ moves the agent forward if $A_{s}$ is smaller than $A_{goal}$.

\begin{figure}
    \centering
    \includegraphics[width=0.5\linewidth]{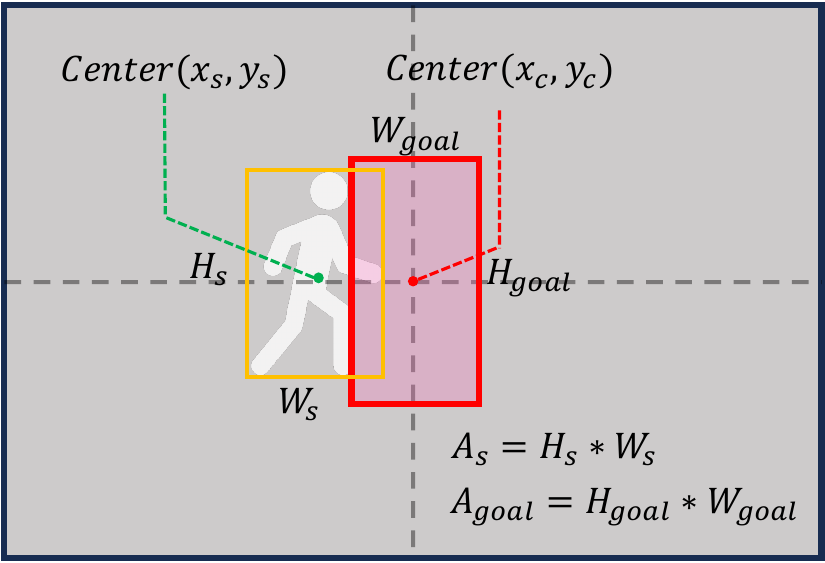}
    \caption{Illustration of the target bounding box (yellow box) and goal bounding box (red box) used in PID controller. The center of target bounding box $Center(x_{s},y_{s})$ and spatial goal $Center(x_{c},y_{c})$ are used to calculate horizontal deviations. The area size of of target bounding box $A_{s}$ and spatial goal $A_{goal}$ are used to calculate distance deviations.}
    \label{fig:PID}
\end{figure}

For angular velocity ($V_{ang}$), the PID controller aims to minimize the angular deviation by computing the difference between the x-axis coordinates of the centers of the target mask $x_{s}$ and the goal mask $x_{c}$, the visualization is shown in Figure ~\ref{fig:PID}. The control input $V_{ang}$ is given by:
\begin{equation}
\begin{aligned}         V_{ang}(t)=
        &K_p (x_{c}-x_{s}(t))+K_i \int_0^t (x_{c}-x_{s}(\tau)) d \tau \\
        &+K_d \frac{d (x_{c}-x_{s}(t))}{d t}
\end{aligned}
\end{equation}where $K_p, K_i$, and $K_d$ are the proportional, integral, and derivative gains, respectively. The angular velocity is constrained within the range ($-30$, $30$).A positive $V_{ang}$ results in a rightward rotation if $x_{s}$ is to the right of $x_{c}$. The final tuned parameters are detailed in the Table ~\ref{tab:bbox_pid}. \\ 
\underline{Ensembled RL:} We trained four separate policy networks corresponding to four representative discrete goals:$[g_{close}, g_{far}, g_{left}, g_{right}]$. The corresponding data for these goals were extracted and filtered from the 1750000 steps offline dataset, providing a focused dataset for training ensemble policies. The policy network architecture is based on the latest state-of-the-art method for Embodied Visual Tracking (EVT) as described by ~\cite{zhong2024empowering}. During evaluation, we choose the corresponding policy based on
the goal spatial position indicated by instruction. \\
\underline{GPT4-o:} We leverage the multi-modal capabilities of GPT4-o to directly generate actions based on the observed image and the desired goal. To ensure smooth and accurate transitions, we developed a system prompt that aids the large model in comprehending the task and regularizes the output format, aligning it with our predefined action settings. This prompt serves as a guiding framework, enabling the model to produce actions that are coherent with the requirements of the task. Specifically, we converted the bounding box goal representation into text-based coordinates, and the input image was first transformed to the same text-conditioned segmentation mask in the paper, then detected the target's bounding box coordinates as input of LLM. Note that the system prompt could directly use image observation as input, but from our experience, the alignment performance is quite poor. The system prompt content is shown by Figure~\ref{prompt_gpt4}. \\

\begin{figure}[h]
    \centering
    \includegraphics[width=0.8\textwidth]{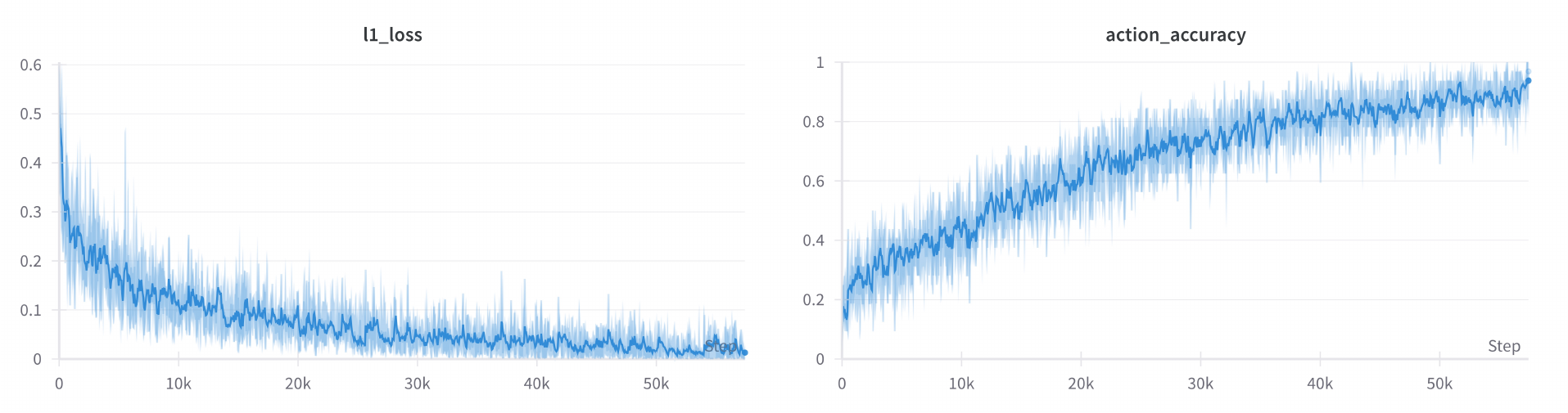}
    \caption{Training curves for OpenVLA fine-tuning on our dataset over 50k steps. The left panel shows the L1 loss, which steadily decreases from an initial value of approximately 0.6 to below 0.05, indicating effective optimization of action prediction. The right panel displays action accuracy, which improves from around 20\% to over 90\%, demonstrating the model's increasing ability to generate appropriate tracking actions given visual observations and language instructions. }
    \label{fig:openvla_training_curve}
\end{figure}

\underline{OpenVLA:} We fine-tuned OpenVLA via LoRA ~\cite{kim2024openvla} using the official implementation and pre-trained model weights from the authors' repository\footnote{Repository URL: \url{https://github.com/openvla/openvla}}. To adapt the model for embodied visual tracking, we modified the action head architecture to output a two-dimensional control signal (linear and angular velocity) compatible with our tracking environment. The fine-tuning process utilized the same trajectory dataset collected for our primary model, comprising 10 million state-action pairs. For each trajectory, we select text instructions from our instruction set based on the corresponding spatial goal of that trajectory. For instance, trajectories with a spatial goal of $(\rho^*, \theta^*) = (200, 0^\circ)$ were paired with instructions such as "Keep the person in the close center." This approach ensured that the fine-tuned model learned the correlation between natural language instructions and appropriate tracking behaviors for various spatial configurations. The detailed hyperparameters we used are listed in Table ~\ref{tab:vla_hyperparameters}. We train the OpenVLA over 50k steps until converge, the training curve are shown in Figure ~\ref{fig:openvla_training_curve}. 

\begin{table}[h]
    \centering
    \caption{Hyperparameters used for fine-tuning OpenVLA}
    \begin{tabular}{ll}
        \toprule
        \textbf{Hyperparameter} & \textbf{Value} \\
        \midrule
        LoRA Rank               & 32 \\
        Batch Size              & 16 \\
        Gradient Accumulation Steps & 1 \\
        Learning Rate           & 5e-4 \\
        Image Augmentation      & True  \\
        \bottomrule
    \end{tabular}
    \label{tab:vla_hyperparameters}
\end{table}

\subsection{Policy Network}
\label{app:policy_network}
In our approach, we use convolutional neural networks (CNN) as the visual extraction module, which is followed by a fully connected layer and a Reward Head. The output of the reward head is used for reward regression during training. The output of the fully connected layer is fed into the recurrent policy network. We use a long-short-term memory(LSTM) network to model temporal consistency.
The recurrent policy network is an extension of the CQL-SAC algorithm~\cite{kumar2020conservative}, where we have modified the data sampling and optimization processes to accommodate the LSTM network. The visual temporal features extracted by the CNN and LSTM are then passed to the actor and critic networks, each consisting of two fully connected layers. The hyperparameters and neural network structures employed in our method are detailed in Table~\ref{HyperParam} and Table ~\ref{cnn}.

\begin{table}[!ht]
\centering
\caption{The hyper-parameters used for offline training and the policy network.}
\begin{tabular}{l|cc}
\hline\hline
Name                 & Symbol           & Value \\ \hline
Learning Rate       & $\alpha$         &  3e-5\\ \hline
Discount Factor     & $\gamma$         &  0.99    \\ \hline
Batch Size           & -         &  128\\ \hline
LSTM update step      & -         & 20 \\ \hline
LSTM Input  Dimension & - & 256 \\ \hline
LSTM Output Dimension & - & 64 \\ \hline
LSTM Hidden Layer size & - & 1 \\ \hline
Reward Head Input Dimension & -& 256 \\ \hline
Reward Head Ouput Dimension & -& 1 \\ \hline
\end{tabular}
\label{HyperParam}
\end{table}

\begin{table}[!t]
    \centering
    \caption{We directly map our adopted action space (continuous actions) from virtual to real. The second and the third columns are the value ranges of velocities in the virtual and the real robot, respectively.}
    \begin{tabular}{c|cc}
    \hline
    \multirow{2}{*}{Bound of Action} & \multicolumn{2}{c}{Linear} \\
    & Virtual (cm/s) & Real (m/s) \\ \hline
    High & 100, 30 & 0.5, 1.0 \\
    Low & -100, -30 & -0.5, -1.0 \\ \hline
    \multirow{2}{*}{} & \multicolumn{2}{c}{Angular} \\
    & Virtual (degree/s) & Real (rad/s) \\ \hline
    High & 100, 30 & 0.5, 1.0 \\
    Low & -100, -30 & -0.5, -1.0 \\ \hline
    \end{tabular}
    \label{tab:action_mapping}
\end{table}

\begin{table*}[!t]
\centering
\caption{The neural network structure, where 8$\times$8-16S4 means 16 filters of size 8$\times$8 and stride 4, FC256 indicates fully connected layer with dimension 256, Reward Head 1 means the fully connected layer for reward regression layer with output dimension 1, and LSTM64 indicates that all the sizes in the LSTM unit are 64.}
\begin{tabular}{c|c|c|c|c|c|c|c}
\hline\hline
Module & \multicolumn{4}{|c|}{Goal-state Aligner} & \multicolumn{3}{c}Recurrent Policy \\ \hline
Layer\# & CNN & CNN & FC & Reward Head & LSTM & FC & FC\\ \hline
Parameters &     8$\times$8-16\emph{S}4     &       4$\times$4-32\emph{S}2       &      256  &1   &       64       &  2 &2 \\
 \hline
\end{tabular}
\label{cnn}

\end{table*}

\section{LLM-based Semantic-Spatial Goal Aligner}
\label{app:instruction parser}
In our method, we develop a hierarchical instruction parser, integrated with Large Language Models (LLM) and chain-of-thought to translate human instructions into a mid-level goal representation. We first describe our evaluation method, then visualize the evaluation result via a radar chart.

\subsection{Quality Evaluation}
To evaluate the effectiveness of our instruction parser and investigate the impact of different reasoning mechanisms on the accuracy of bounding box generation, we employ three different methods: \underline{GPT-4o}: We use GPT-4o with a system prompt that includes both a task introduction and chain-of-thought (CoT) reasoning guidelines for evaluation. \underline{GPT-4o w/o CoT}: We use GPT-4o with a system prompt containing only the task introduction for evaluation. \underline{GPT-o1}: We use GPT-o1 with a system prompt that includes the task introduction for evaluation.  

We ensure a fair evaluation by sampling 140 instructions from the instruction list in Table ~\ref{tab:instructions}, and each instruction is paired with a corresponding spatial goal. This setup minimizes ambiguity in the textual instructions, providing a baseline for evaluating the correctness of the generated bounding boxes. The performance is assessed by the accuracy of generated bounding boxes. The accuracy is calculated as: the number of correctly generated bounding boxes divided by the total sampled instances

For each instruction, we define two categories based on the type of spatial instruction: absolute spatial positions and relative spatial position changes. The rules for evaluating bounding box correctness are as follows:

1)Instructions conveying absolute spatial positions (first 22 rows of Table 4): We use the final generated bounding box $[x,y,w,h]$, calculate horizontal position $x$ and area size $w*h$ to determine correctness.
\begin{itemize}
    \item{ $(\rho^*, \theta^*) = (200, 0^\circ)$: Valid if area size is within $(0.06, 0.3)$ and $x$ is within $(0.4, 0.6)$.}
    \item{$(\rho^*, \theta^*) = (450, 0^\circ)$: Valid if area size is within $(0, 0.06)$ and $x$ is within $(0.4, 0.6)$}
    \item{$(\rho^*, \theta^*) = (350, -20^\circ)$: Valid if $x$ is within $(0, 0.4)$.}
    \item{$(\rho^*, \theta^*) = (350, 20^\circ)$: Valid if $x$ is within $(0.6, 1)$.}
\end{itemize}

2)Instructions conveying relative spatial position change (last 20 rows of Table 4): We evaluated based on bounding box increments $[\Delta x,\Delta y, \Delta w, \Delta h]$ generated by the parser.
\begin{itemize}

\item{ $(\Delta \rho, \Delta \theta) = (-150, 0^\circ)$: Valid if both $\Delta w$ and $\Delta h$ are positive.}
\item{$(\Delta \rho, \Delta \theta) = (150, 0^\circ)$: Valid if both $\Delta w$ and $\Delta h$ are negative.}
\item{$(\Delta \rho, \Delta \theta) = (0, -20^\circ)$: Valid if $\Delta x < 0$.}
\item{$(\Delta \rho, \Delta \theta) = (0, 20^\circ)$: Valid if $\Delta x > 0$.}
\end{itemize}
\begin{figure}[!t]
    \centering
    \includegraphics[width=\linewidth]{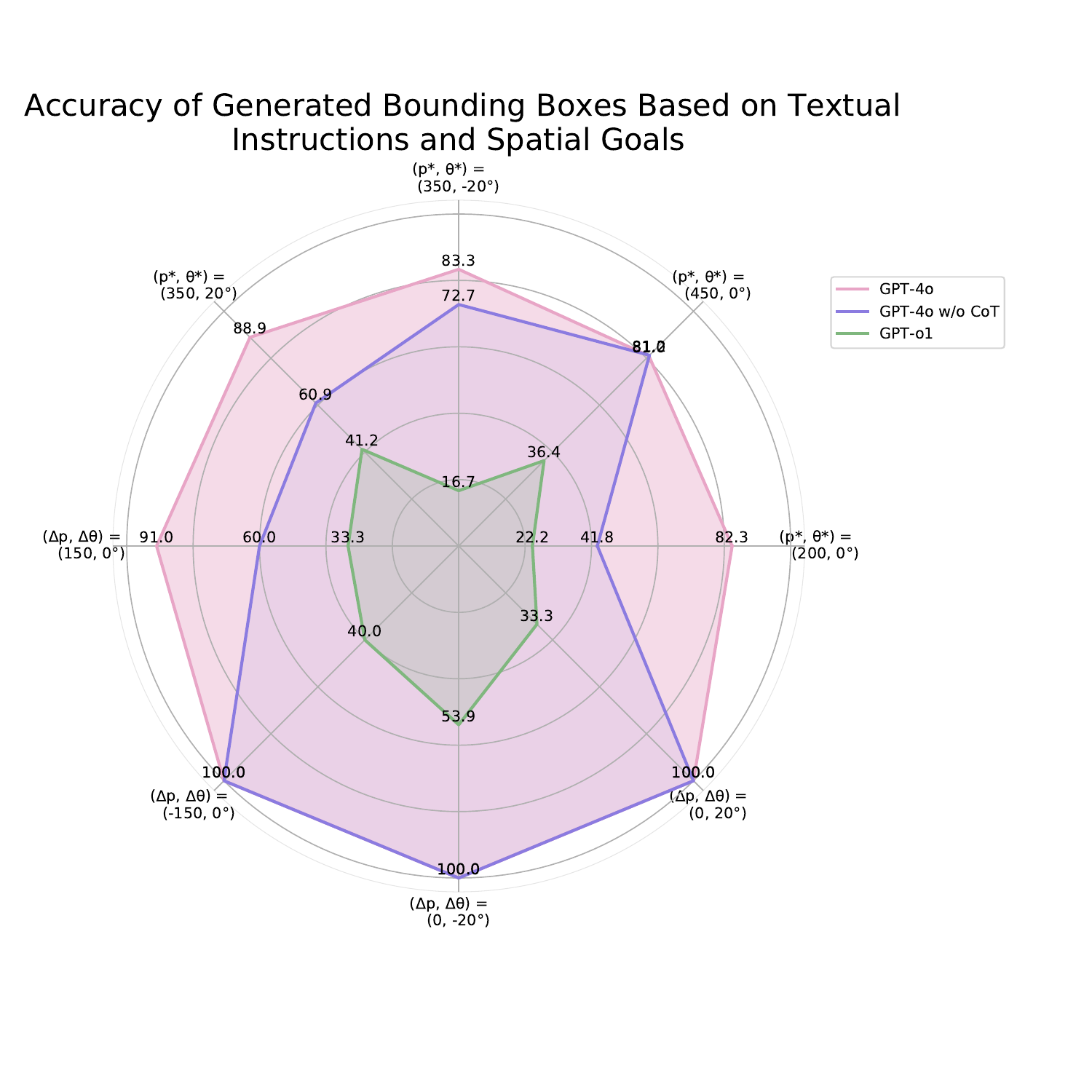}
    \vspace{-1cm}
    \caption{The accuracy of generated bounding boxed based on textual instructions and spatial goals by using GPT-4o, GPT-4o w/o CoT, and GPT-o1.}
    \label{fig:eval_parser}
\end{figure}
The results in Figure ~\ref{fig:eval_parser} confirm that the chain-of-thought (CoT) reasoning mechanism significantly improves the accuracy of the bounding box generation. Compared with GPT-4o w/o CoT and GPT-o1,  GPT-4o consistently exceeds 80\%, achieving up to 100\% accuracy in some cases, demonstrating its reliability in translating textual instructions into spatially accurate bounding boxes. We argue that CoT enhances the model’s ability to handle more nuanced spatial relationships. GPT-O1 performs significantly worse than GPT-4o and GPT-4o w/o CoT, and we believe this is largely due to the underlying reasoning mechanism. GPT-o1 employs an automatic decomposition reasoning approach, which decomposes tasks into smaller steps without considering the broader context. This lack of holistic reasoning leads to poor performance in our task, particularly when handling complex spatial relationships.

\subsection{Prompts}
\label{app:prompt}
We define a system prompt aiming to help the LLM understand the tracking task and introduce the Chain-of-thought (CoT) to enhance the LLM's understanding ability. The detailed content of the system prompt is shown in Figure~\ref{prompt_instruction}.
\section{Graphic User Interface}
\begin{figure}
    \centering
    \includegraphics[width=0.6\linewidth]{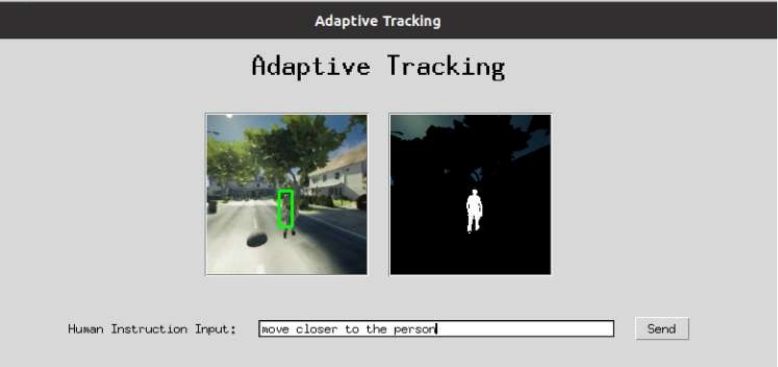}
    \caption{A real-time interactive GUI snapshot. The left panel displays the tracker's first-person view, while the right panel shows the text-conditioned mask generated by the game engine. Users can input instructions in the text box at the bottom or draw a bounding box (e.g., the green box in the left image) and click the "Send" button to interact with the system.}
    \label{fig:GUI}
\end{figure}

To enable an intuitive visual interaction and multi-modal instruction input, we design a simple GUI for user input instructions while observing the environment from the tracking agent's first-person view. Users could directly type text instructions and click the ``send" button to update instructions. The GUI demonstration is shown in Figure ~\ref{fig:GUI}. 
\begin{figure}[h]
    \centering
    \includegraphics[width=0.5\linewidth]{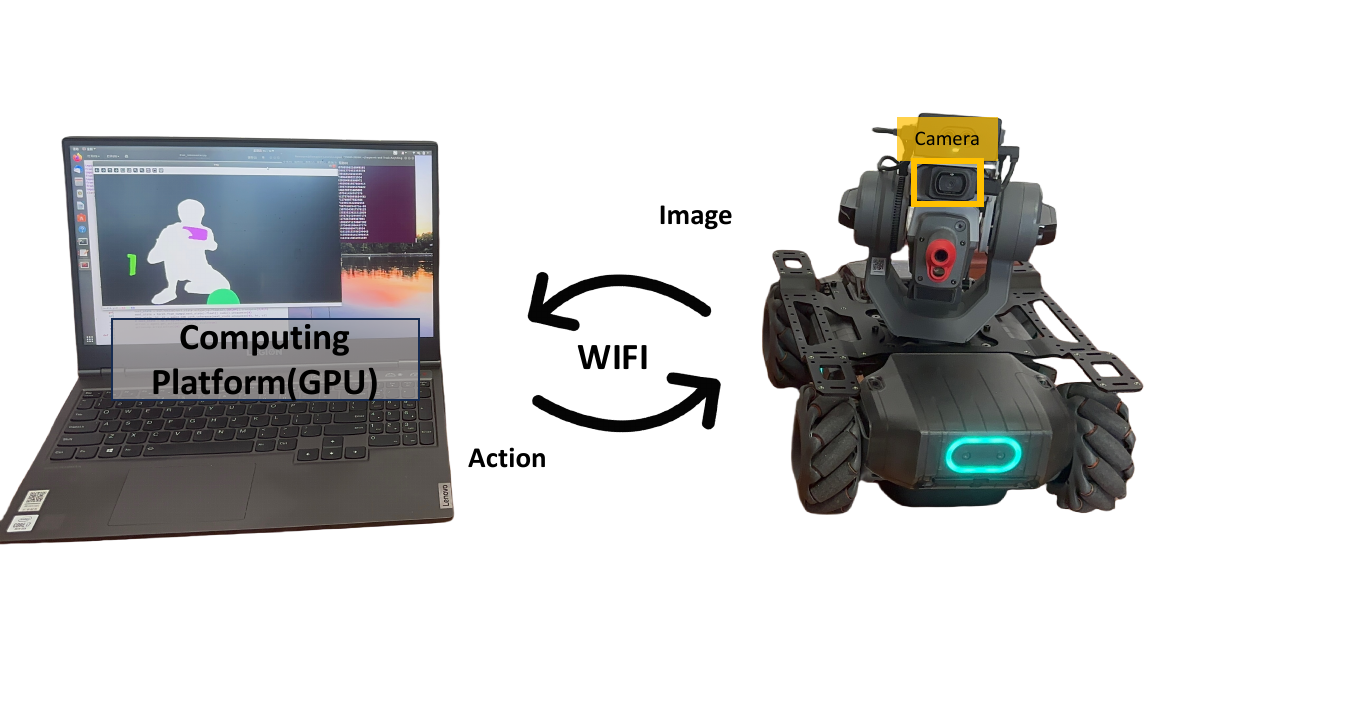}
    \caption{In this real-world deployment, the robot's onboard camera captures visual data, which is wirelessly transmitted to a laptop for real-time processing. The model on the laptop interprets the incoming images, generates appropriate action commands, and transmits them back to the robot via WiFi, enabling precise control of its movements.}
    \label{fig:robot}
\end{figure}

\section{Real-world Deployment}
We transfer our agent into real-world scenarios to verify the practical contribution and the effectiveness of our proposed method. Comprehensive video demonstrations of these experiments are available on our project website (https://sites.google.com/view/hievt).

\label{app:hardware}
\subsection{Hardware setup}
To evaluate our proposed method in real-world scenarios, we use RoboMaster EP \footnote{https://www.dji-robomaster.com/robomaster-ep.html} a 4-wheeled robot manufactured by DJI, as our experimental platform (Figure ~\ref{fig:robot}). Equipped with an RGB camera, the RoboMaster captures images and allows direct control of linear speed and angular motion via a Python API. This advanced design ensures precise control over the robot's movements during our experiments. To enable real-time image processing, we use a wireless LAN connection to transmit camera data to a laptop with an Nvidia RTX A3000 GPU, which acts as the computational platform to run the model and predict actions based on raw pixel images. The corresponding control signals are then sent back to the robot, completing a closed-loop control system. This seamless integration of hardware and software allows us to execute complex tasks efficiently and accurately, advancing sim-to-real experimentation. In the training phase, we use a continuous action space, which enables us to directly map the speed to robot control signals. The mapping relationship is shown in Table ~\ref{tab:action_mapping}.

\subsection{Experimental Results and Observations}

The real-world experiments were conducted to verify three key hypotheses: (1) the system can maintain robustness across diverse physical scenarios, (2) our hierarchical framework can effectively bridge the gap between language instructions and tracking behavior in real environments,  and (3) the intermediate spatial goal representation provides sufficient flexibility for diverse instruction types. 

\subsubsection{Adaptive Response to Dynamic Movement}
The robot successfully maintained tracking while targets performed unpredictable movements including \textbf{sudden direction changes} and \textbf{varying speeds( from walking to running).} A critical observation from these experiments was that the system's performance directly validated our hierarchical design approach. The instruction reasoning process (running at approximately 1-2 Hz) did not interfere with the high-frequency control policy, allowing the tracking policy to operate continuously even during instruction processing. This asynchronous execution enabled the system to maintain responsive tracking while simultaneously processing complex instructions—a capability that would be impossible with end-to-end approaches like GPT-4o or OpenVLA where inference latency directly impacts control frequency. Our video demonstrations clearly show this advantage in action: even when new instructions are being processed (visible through UI indicators in the videos), the robot maintains smooth, continuous tracking without pauses or hesitation. This confirms that our hierarchical framework effectively decouples semantic understanding from real-time control, enabling robust performance in dynamic real-world scenarios.
\subsubsection{Instruction Understanding}
Our system demonstrated sophisticated spatial reasoning capabilities through its response to various instructions:
\begin{itemize}
    \item \textbf{``Move closer to the target'':} The robot accurately adjusted its position to reduce distance while maintaining the target in view. Video evidence shows the system precisely approaching stationary persons and maintaining desired relative positions, continuously tracking while preserving spatial alignment.

    \item \textbf{``Get closer to the target and keep him in the left'':} When instructed to reduce distance while maintaining left positioning, the system generated a larger bounding box on the left and swiftly adjusted its relative position accordingly. Our videos document smooth transitions between different instructions, highlighting the system's adaptability.

    \item \textbf{``Move further from the target and keep him in the left'':} The robot successfully interpreted both distance and positioning components of this instruction, generating an appropriate smaller bounding box positioned on the left side of the frame under different initial positions. Two video demonstrations clearly exhibit consistent left-oriented positioning while maintaining increased distance.
\end{itemize}

\subsubsection{Spatial Goal Adaptation}
These visual recordings confirm that our intermediate spatial goal representation effectively bridges the gap between natural language instructions and robot behavior. Our adaptive policy could respond to dynamically changed spatial goals fast and accurately.

\begin{figure*}[htbp]
  \centering
\begin{tcolorbox}[title=System Prompt]
\begin{lstlisting}[texcl=true, escapechar=|]
Objective: 
You are an intelligent tracking agent designed to follow human instructions and dynamically adjust your tracking goal between you and the target. Your task is twofold:
1.Identify the target category mentioned in the instruction (e.g., "person," "vehicle," "object").
2.Understand the human instruction to determine the tracking goal. The goal is represented as an expected bounding box position in your field of view, with the center at [cx, cy], width w, and height h. This will guide your tracking strategy to align the target object with the specified bounding box. The local control strategy will then use this expected bounding box to achieve different tracking angles and distances based on human instruction.

Representation detail:
All positions in the task should be represented as normalized bounding box coordinates relative to the image size in the field of view with width and height (e.g., [cx, cy, w, h]). 'cx' and 'cy' represent the center of the bounding box, and 'w' and 'h' represent the width and height of the bounding box, respectively, all normalized to the range [0, 1].

Task Understanding:
1. **Instruction:** A natural language command describing the desired change in the tracking of the target (e.g., "Get closer to the person," "Move further from the car," "Keep the dog in the center," or "Keep the object on the left").
2. **Current bounding box:** The current bounding box coordinates and size of the target in your field of view relative to the image size, normalized to [0, 1] (e.g., "Target position: [cx, cy, w, h]").

Task Definition:
Your task is to:
Extract the target category from the instruction (e.g., "person," "car," "dog") and determine the expected bounding box position and size within your field of view based on the human instruction and the current target position.

This should include:
1.**Target category:** Based on the human instruction, provide the target category name.
2.**Bounding Box Increment:** Based on the human instruction, provide the change in the bounding box. This should be represented as [|$\Delta$|cx, |$\Delta$|cy, |$\Delta$|w, |$\Delta$|h], where |$\Delta$|cx and |$\Delta$|cy are the changes in the center coordinates, and |$\Delta$|w and |$\Delta$|h are the changes in the width and height of the bounding box.

Instructions: Given the provided human instruction, and current position, think step by step, decide the target category name and best bounding box increment to meet the human's instructions.

Strategy Considerations:
The given Current Position represents the current target distance and angle relative to the tracker.
The human instruction represented the demand for expected tracking distance and angle between the target and tracker.
You should consider the human instruction first, and transform the abstract instruction into the concrete bounding box position increment.
The increment value should be 20% each time with respect to the original proportion of the target bounding box.
The increment value should consider the first-person view perspective effect.

Provide the target category name format in "**Target category:** [target category]" and the chain-of-though process of the increment of bounding box position and size end with the format "**Bounding Box Increment:** chain-of-thought process.[|$\Delta$|cx, |$\Delta$|cy, |$\Delta$|w, |$\Delta$|h]" in [output:]

Example:
Example:
[input:]
Instruction: "Get closer to the person."
Current bounding box: [0.51, 0.57, 0.07, 0.14].

[output:]
**Target category:** [person]
**Bounding Box Increment:** The current bounding box indicates it is positioned near the center of view. If the instruction wants to get closer to the target, the bounding box size should be larger without horizontal change and a slight increment in vertical position, which should be increased |$\Delta$|w, |$\Delta$|h and |$\Delta$|cy.[0.0, 0.05, 0.03, 0.17].

\end{lstlisting}
\end{tcolorbox}
  \caption{System prompt used in instruction parser module.}
  \label{prompt_instruction}
\end{figure*}

\begin{figure*}[htbp]
\centering\begin{tcolorbox}[title=System Prompt]
\begin{lstlisting}
You are an intelligent tracking agent designed to generate discrete control actions to control the robot to follow the target object at an expected distance and angle.

Representation detail:
Bounding box: The bounding box position and size within your field of view, represented as [cx, cy, w, h]. 'cx' and 'cy' represent the center of the bounding box, and 'w' and 'h' represent the width and height of the bounding box, respectively.

Control Actions:  The control actions are discrete and include the following:
    -move forward: control the robot to move forward for 0.5 meters.
    -move backward: control the robot to move backward for 0.5 meters.
    -stop: control the robot to stop moving.
    -turn left: control the robot to move forward for 0.25 meters and turn left for 15 degrees.
    -turn right: control the robot to move forward for 0.25 meters and turn right for 15 degrees.
    
Task Understanding:
1. **Goal bounding box:** This is provided by the user to indicate the expected distance and angle between the target and the tracker, which is a bounding box format, the agent should try to align the target bounding box with the goal bounding box as much as possible.
2. **Target bounding box:** This is provided by user to indicate the current target position in the image, in the bounding box format.

Task Definition:
Your task is to give a suitable action from Control actions, and try to align the Goal bounding box with the target bounding box as much as possible.
1. **Actions:** Based on the given Goal bounding box and Target bounding box, you should provide the best control action from the control actions to align the target bounding box with the goal bounding box as much as possible.

Strategy Considerations:
The target bounding box size in the image represents the spatial distance, the smaller size corresponds to a larger distance between the robot to the target.
If the center of the target bounding box is on the right side of the goal bounding box center, the robot should turn right to align the target bounding box with the goal bounding box. In contrast, if the center of the target bounding box is on the left side of the goal bounding box center, the robot should turn left to align the target bounding box with the goal bounding box.

Instructions: Given the provided target bounding box and goal bounding box, decide the best action to adjust the robot's position.
Provide ONLY and the increment of bounding box position and size in [output:] format in [Control Action] without additional explanations or additional text.

\end{lstlisting}
\end{tcolorbox}
  \caption{System prompt used in baseline method GPT4-o. }
  \label{prompt_gpt4}
\end{figure*}



\end{document}